\documentclass[]{LeafLabLaTexTemplate}
\input{preamble}

\newcolumntype{Y}{>{\centering\arraybackslash}X}
\newif\ifaddtitlelogo

\addtitlelogofalse

\usepackage{enumitem}   % \begin{enumerate}[label=Q\arabic*:, leftmargin=10mm]
\usepackage{nicefrac}   % \nicefrac{alt}{2} in the Meta-World true rewards
\usepackage{float}      % [H] float placement
\usepackage{wrapfig}    % figures that were one IEEE column wide now wrap text
\newcolumntype{T}{>{\scriptsize}c}

\crefname{section}{Section}{Sections}
\crefname{subsection}{Section}{Sections}
\crefname{equation}{Equation}{Equations}
\crefname{figure}{Figure}{Figures}
\crefname{table}{Table}{Tables}
\crefname{algorithm}{Algorithm}{Algorithms}
\crefname{appendix}{Appendix}{Appendices}

\title{Residual Reward Models: Leveraging Prior Knowledge for Efficient
Preference-based Reinforcement Learning in Robotics}

\author[1,*]{Chenyang Cao}
\author[1]{Miguel Rogel-García}
\author[1]{Mohamed Nabail}
\author[2]{Xueqian Wang}
\author[1]{Nicholas Rhinehart}

\affiliation[1]{Robotics Institute, University of Toronto}
\affiliation[2]{Tsinghua University}

\abstract{
Preference-based Reinforcement Learning (PbRL) provides a promising alternative to heuristic reward design in complex robotic environments. However, PbRL often suffers from poor sample efficiency, requiring extensive and costly human feedback, which limits its real-world applicability. Prior work has proposed learning a reward model from demonstrations and fine-tuning it using preferences. However, when the model is a neural network, transitioning between different loss functions across training phases often leads to unstable optimization and performance degradation. In this paper, we propose a method to effectively leverage prior knowledge with a Residual Reward Model (RRM). An RRM assumes that the true reward of the environment can be split into a sum of two parts: a prior reward and a learned reward. The prior reward is a term available before training, such as an engineering heuristic ``best guess'', a language-generated reward, or a reward function learned from inverse reinforcement learning, and the learned reward is then trained with preferences as a residual offset. Experimental results in Meta-World and DM-Control show that RRMs substantially improve the sample efficiency of common PbRL methods across various prior reward types. Furthermore, we demonstrate the practical efficacy of our method on a physical Franka Panda robot, accelerating policy learning and achieving high success rates in fewer steps than baselines.
}

\metadata[
\raisebox{-0.35em}{\includegraphics[width=0.05\linewidth]{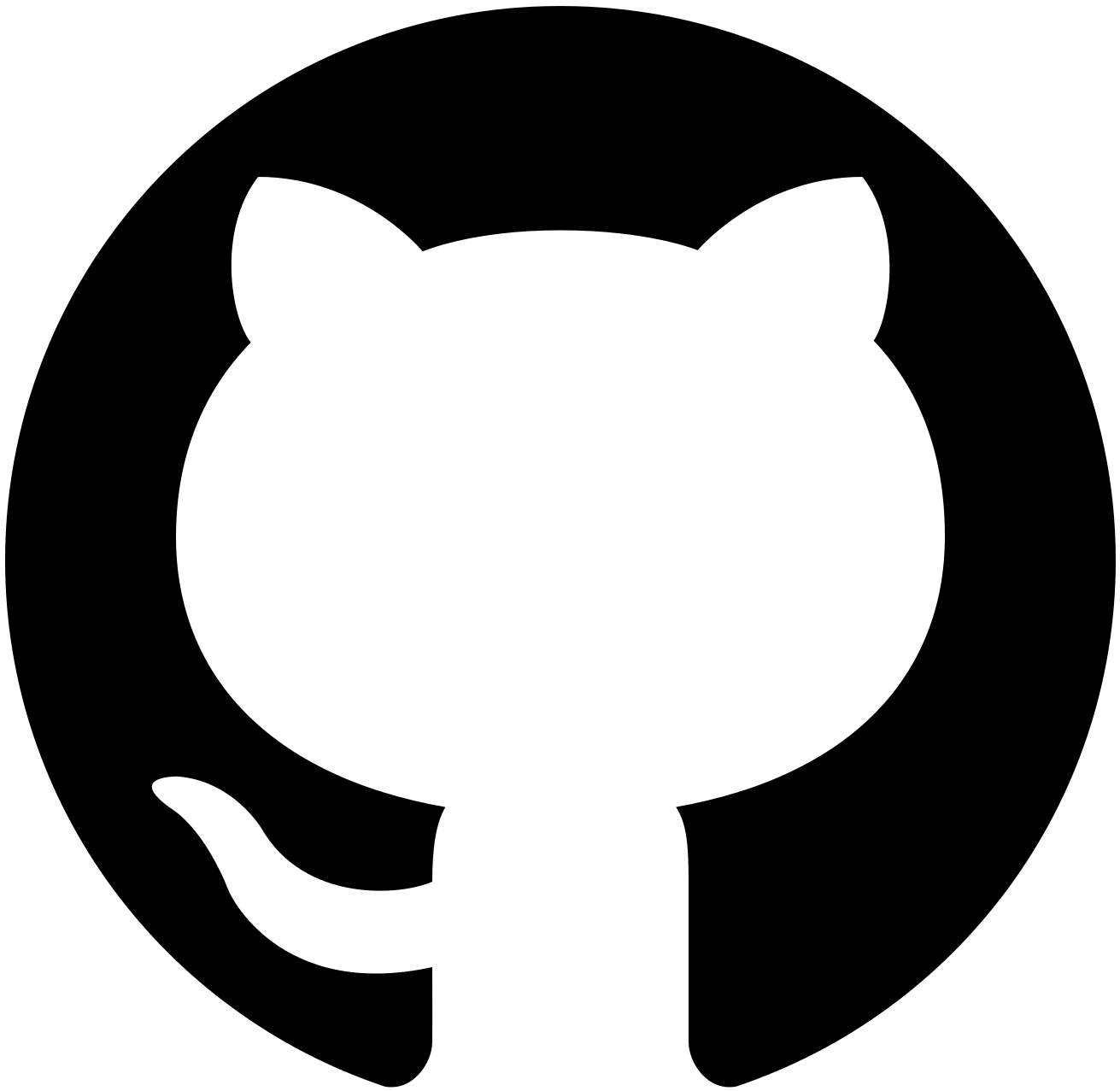}}~~ Project Website]{\href{https://sunlighted.github.io/RRM-web/}{\textbf{https://sunlighted.github.io/RRM-web/}}
}
\begin{document}

\maketitle

% Unnumbered footnote at the bottom of the title page (\blfootnote is defined in the .cls).
\blfootnote{%
  $^{*}$Correspondence to Chenyang Cao, \email{chenyang.cao@mail.utoronto.ca}.%
}

\section{Introduction}
\label{sec:intro}

\begin{wrapfigure}[15]{r}{0.38\textwidth}
  \centering
  \vspace{-14pt}
    \includegraphics[width=\linewidth]{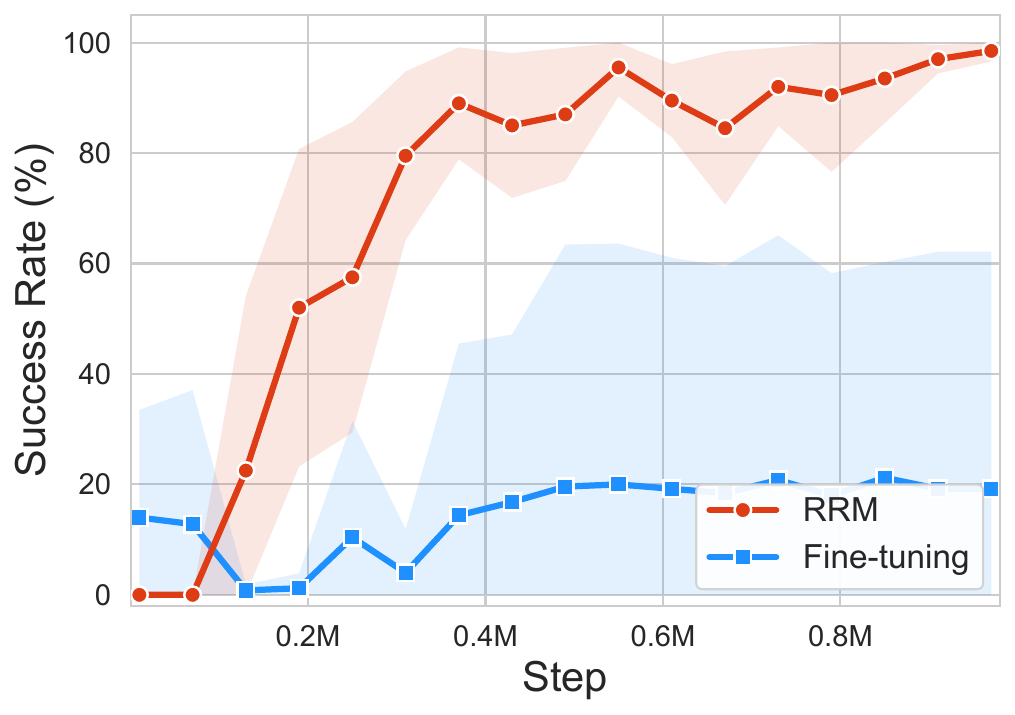}
    \caption{\textbf{Fine-tuning vs.\ RRM} on \textit{Button-press}. Fine-tuning an imitation reward plateaus at $20\%$ success; RRM reuses it as an additive prior and reaches $100\%$.}
    \label{fig:ftmain}
\end{wrapfigure}

Reward is a crucial component in Reinforcement Learning (RL), yet designing well-aligned reward functions for robots to acquire complex physical skills is usually difficult \cite{singh2009rewards, abel2021expressivity}. Reward function designers are constantly confronted with specifying necessary kinematic attributes and handling unforeseen physical scenarios \cite{hadfieldmenell2020inverserewarddesign}. Furthermore, uninstrumented real-world environments lack the state access required for dense reward formulation, increasing the risk of suboptimal or unsafe robot behaviors \cite{zhu2020ingredientsrealworldroboticreinforcement}. Reward learning offers a promising data-driven alternative, learning a reward function directly from environmental or human-provided alignment signals to produce a near-optimal policy for the implicit robotic task.

A popular approach for this and the focus of our study is Preference-based Reinforcement Learning (PbRL). PbRL facilitates the automatic acquisition of reward functions from human feedback, mitigating the burden of manual reward engineering \cite{lee2021bpref}. During training, PbRL aligns the reward model by querying a human expert for preferences over pairs of trajectory segments. Therefore, PbRL can improve the performance of RL on highly complex tasks by learning more desirable behaviors while avoiding reward exploitation \cite{christiano2017pbrl, lee2021pebblefeedbackefficientinteractivereinforcement, park2022surf, liang2022rune}. However, PbRL is heavily constrained by feedback efficiency and convergence speed; requiring a human operator to provide a large amount of feedback is too expensive for real-world robotic training \cite{christiano2017pbrl}. An alternative to PbRL, Inverse Reinforcement Learning (IRL), estimates a reward from expert demonstrations \cite{abbeel2004apprenticeship}. Yet, collecting high-quality teleoperated demonstrations is expensive and not necessarily available for complex behaviors \cite{liu2023learning}. Other mitigation methods include physical corrections \cite{li2021learninghumanobjectivessequences}, natural language instructions \cite{coreyes2019guidingpolicieslanguagemetalearning}, disengagements \cite{kahn2020landlearningnavigatedisengagements}, or interventions \cite{korkmaz2025milemodelbasedinterventionlearning}. However, these approaches often require specialized hardware interfaces, complex language grounding, or struggle to generalize across diverse tasks.

We aim to design a highly sample-efficient PbRL method that incorporates general sources of prior knowledge as ``reward priors''. While one approach pre-trains a reward model via IRL and fine-tunes it by PbRL \cite{palan2019demopref, biyik2022demopref2}, this method is highly vulnerable to distribution shift between the static dataset and the deployment environment, causing the reward to fail to converge. Additionally, sequentially training a neural-network reward model under different objectives often destabilizes optimization: the abrupt change in training signal causes the fine-tuned reward to degrade rather than improve. We illustrate this instability in \cref{fig:ftmain}, where fine-tuning stalls well below RRM built on the same imitation reward.

%—such as using MSE loss on demonstrations followed by cross-entropy loss on preference data—

\begin{figure}[t]
  \centering
  \includegraphics[width=1.0\textwidth]{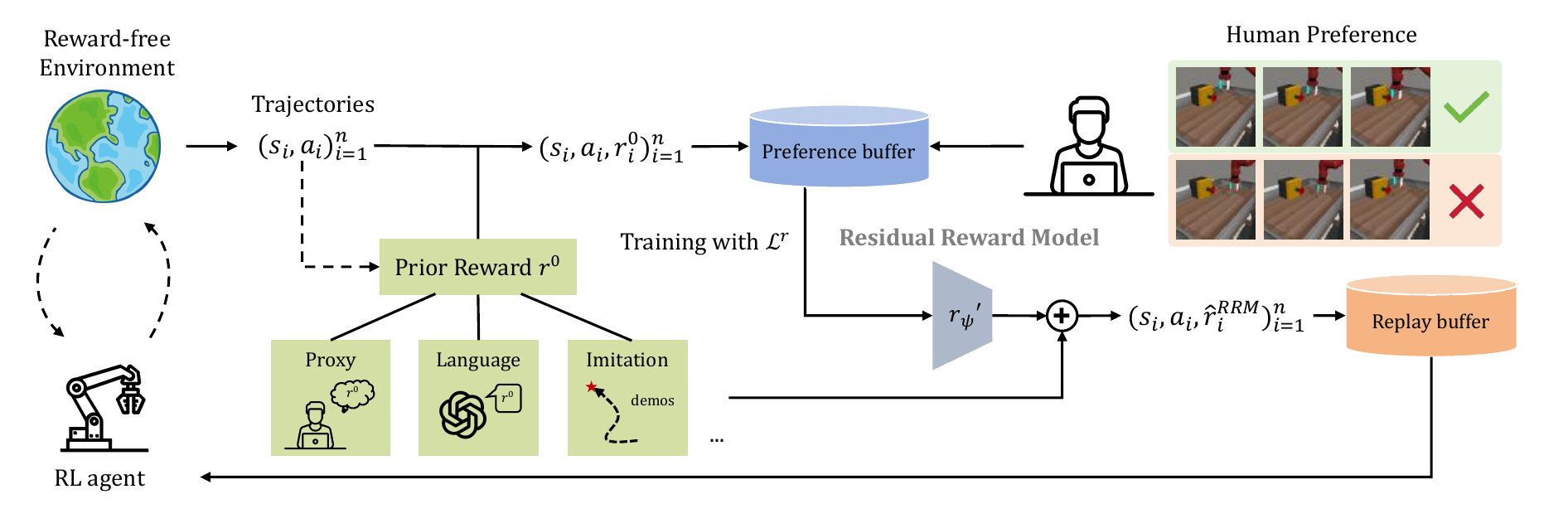}
  \caption{\textbf{Residual Reward Model.} An agent interacts with the reward-free environment and generates trajectories. In order to obtain rewards for reinforcement learning, our method assumes access to a ``prior'' reward that conveys some information about the task, but generally may be different from the true task reward function. These prior rewards form part of a reward function that is trained to be aligned with preference pairs.}
  \label{fig:RRM}
\end{figure}

We draw inspiration from the structure of residual networks \cite{he2016identity, johannink2019residual} and apply it to PbRL by adding a preference-based reward model as a residual term to the prior reward. The prior reward refers to a reward function known at the start of training, such as a human's ``best guess'' rewards, language-generated (e.g., Large Language Model) rewards, or rewards obtained from a reward learning method (e.g., IRL). The residual reward outputs an offset to correct the prior where it contradicts human preferences, leaving the underlying prior reward unchanged. We call this structure a Residual Reward Model (RRM), as shown in \cref{fig:RRM}. The presence of the prior rewards enables RRM to ground the robot's exploration in goal-directed behaviors early in training, making it easier for human labelers to evaluate trajectory segments. At the same time, the residual reward introduces corrections for higher-level task alignment. Consequently, the structure design not only improves the performance of PbRL algorithms but also reduces the absolute volume of required feedback. Moreover, the prior reward serves as a stabilizing anchor, ensuring robust policy learning even when encountering noisy or irrational human feedback.

To validate our approach, we construct manually defined rewards, LLM-generated rewards, and imitation learning-based rewards for RRMs and test their performance across simulated manipulation tasks (both state-based and image-based) in Meta-World \cite{yu2020meta} and state-based locomotion tasks in DM-Control \cite{tunyasuvunakool2020}. Our method consistently outperforms baselines in sample efficiency and convergence speed, particularly under restricted or noisy feedback settings, while exhibiting robustness to semantically negated priors. Finally, we demonstrate the efficacy of our method through zero-shot sim-to-real transfer, evaluating the policy directly on a physical Franka Panda robot. The results demonstrate that RRMs significantly accelerate task learning, reaching high success rates in real-world manipulation faster than the baseline, improving the robot learning efficiency.

The main contributions of this work are as follows:
\begin{enumerate}[leftmargin=10mm]
    \item We introduce Residual Reward Models (RRMs), a reward learning framework that takes advantage of prior knowledge. We show RRMs can be used to augment the performance of existing PbRL methods, focusing on PEBBLE \cite{lee2021pebblefeedbackefficientinteractivereinforcement}, as well as SURF \cite{park2022surf} and MRN \cite{liu2022mrn} by achieving higher sample efficiency. 
    \item RRMs show robustness to typical constraints in physical deployment, including limited total feedback, low feedback frequency, and noisy or even semantically negated prior rewards.
    \item We show that RRMs lead to high success rates on a real robot faster than the baseline method (upon which RRMs are built) in three sets of manipulation experiments.
\end{enumerate}

\section{Related Work}
\label{sec:relat}

\subsection{Residual Learning in RL}

Residual policy learning was introduced to improve robotic tasks by building on an imperfect controller rather than learning a policy from scratch. This approach improves data efficiency and addresses intractability challenges \cite{silver2019residualpolicylearning, johannink2019residual}. In the context of human-in-the-loop RL, this has been further extended to first learn a base policy from demonstrations, and then use RL to learn a residual policy that corrects its actions \cite{alakuijala2021residualreinforcementlearningdemonstrations}. Prior work has demonstrated that residual reinforcement learning can enhance robot control by integrating learned policies with residual corrections  \cite{haldar2023teach,ankile2024fromimitation}. Different from them, we consider using residual learning in reward modeling.

\subsection{Reward Learning from Human Preferences}

Learning reward functions and RL policies from pairwise trajectory preferences is a widely used approach since it is often easier for people to rank behaviors rather than demonstrate an ideal one and easier to translate to reward updates as opposed to physical corrections and language instructions \cite{hejna2022fewshotpreferencelearninghumanintheloop}. Our work mainly builds on PEBBLE \cite{lee2021pebblefeedbackefficientinteractivereinforcement}, and can be extended to other PbRL algorithms \cite{park2022surf, liu2022mrn}.

A major challenge in preference learning is achieving sample efficiency, as the need for frequent human queries causes significant bottlenecks for real-world robotic experiments \cite{lee2021bpref, park2022surf, liang2022rune}. While prior work improves efficiency by sampling informative behaviors \cite{lee2021pebblefeedbackefficientinteractivereinforcement, hu2023qpm, biyik2018active} or leveraging meta-learning and mixed feedback \cite{palan2019demopref, hejna2022fewshotpreferencelearninghumanintheloop, pamies2023autonomous}, these methods often cannot directly guide the agent toward the target task or assume reusable past feedback \cite{NEURIPS2020_2f10c157, kumar2020discor, jiang2024proxy}. Closely related to our work is the integration of baseline objectives with preferences \cite{marta2023aligning}. However, it treats the baseline reward as a separate, competing objective, requiring manually tuned weights to find a Pareto trade-off. In contrast, our method explicitly treats the prior reward as an initial guess and learns a residual term to directly correct formulation errors. This eliminates tedious hyperparameter tuning and actively leverages prior knowledge to accelerate policy convergence, making it highly practical for physical robot deployments.

\section{Residual Reward Modeling}
\label{sec:method}

In this section, we present our method, Residual Reward Models (RRMs), by describing the main components -- the sources of the prior reward and how the residual reward is learned from Preference-based Reinforcement Learning (PbRL). We now formalize the problem setting.

\subsection{Reinforcement Learning (RL)} We use a standard Markov Decision Process (MDP) formulation $\mathcal{M} = (\mathcal{S}, \mathcal{A}, \mathbb{P}, r, \mu, \gamma)$ with discrete time steps $t \in \{0,\ldots, T \}$. $\mathcal{S}$ and $\mathcal{A}$ denote the state and action space. $\mathbb{P}(s'|s,a)$ represents the transition probability from $s$ to $s'$ under action $a$. $r: \mathcal{S} \times \mathcal{A} \rightarrow \mathbb{R}$ is reward function. $\mu(\cdot): \mathcal{S}\rightarrow[0,1]$ is the initial state distribution and $\gamma\in (0,1)$ is the discount factor. 

\subsection{Preference-based Reinforcement Learning (PbRL)} Instead of getting reward signals from the environment in a common RL setting, PbRL assumes $r$ is unknown and unsampleable. Following previous work, we consider a basic framework that learns a reward function from human preferences. The learned reward function is $\hat{r}_\psi: \mathcal{S} \times \mathcal{A} \rightarrow \mathbb{R}$. During training, a trajectory segment $\sigma_i = \{s^i_t, a^i_t\}_{t=0}^{H-1}$ can be observed by human, where $H$ is much shorter than the episode length. Humans can receive a pair of segments $(\sigma_0, \sigma_1)$ at some specific training steps and provide a preference $y \in \{(0,1), (1,0)\}$ to indicate which segment is better. The preference label $y = (1,0)$ means $\sigma_0$ is preferred to $\sigma_1$, i.e., $\sigma_0 \succ \sigma_1$, and $y = (0,1)$ otherwise. A preference buffer $\mathcal{D}$ is built by storing the history preferences as triples $(\sigma_0, \sigma_1, y)$. The Bradley-Terry model \cite{bradley1952rank} is used to model the preference predictor by using the reward function $\hat{r}_\psi$, as shown in \cref{eq:btmodel}:
\begin{equation}\label{eq:btmodel}
    P_\psi[\sigma_0 \succ \sigma_1] = \frac{\exp(\sum_{t=0}^{H-1}\hat{r}_\psi(s_t^0, a_t^0))}{\sum_{i\in\{0,1\}}\exp(\sum_{t=0}^{H-1}\hat{r}_\psi(s_t^i, a_t^i))}.
\end{equation}
The reward model $\hat{r}_\psi$ is optimized by minimizing the cross-entropy loss shown in \cref{eq:btloss}:
\begin{equation}\label{eq:btloss}
\begin{aligned}
    \mathcal{L}^{r} =& -\mathop{\mathbb{E}}\limits_{{(\sigma_0, \sigma_1, y)\in \mathcal{D}}} \Big[y(0)\log P_\psi(\sigma_0 \succ \sigma_1) + y(1)\log P_\psi(\sigma_1 \succ \sigma_0) \Big].
\end{aligned}
\end{equation}
We now describe Residual Reward Models (RRMs) for PbRL. RRMs assume that the true reward of the task can be divided into two components: a prior $r^0$ and a learned residual $r'_{\psi}$.
\begin{equation} \label{eq:rrm}
    \hat{r}^{\text{RRM}}_\psi(s,a) = r^0(s,a) + r'_{\psi}(s,a, r^0(s,a)).
\end{equation}
Note that \cref{eq:btmodel,eq:btloss} describe the general preference learning framework. In our RRM setting, we substitute the general reward $\hat{r}_\psi$ in these equations with $\hat{r}^{\text{RRM}}_\psi$. Crucially, the residual network $r'_{\psi}$ is explicitly conditioned on the prior knowledge by taking the prior reward value $r^0(s,a)$ as an additional input feature alongside the state-action pair. This design shares the core philosophy of residual learning architectures like ResNet \cite{he2016identity}.

% This relationship is straightforward to motivate with Bayes' rule. The prior reward $r^0$ refers to the reward information available at the start of training. It represents the ``best guess'' for a suitable reward function, and can be hand-crafted, generated by LLM, obtained from IRL, or from other reward learning methods. In contrast, the learned reward $r'_{\psi}$ is obtained by adjusting $r^{\text{RRM}}_\psi$ to be aligned with preference data (as measured by Eq.~\ref{eq:btloss}). It essentially outputs the necessary offset to make the prior reward consistent with human feedback, and can correspond to both small (e.g., at state-action pairs where small or no residual is needed) and large adjustments.

We estimate $r'_{\psi}$ via Maximum A Posteriori (MAP) inference, seeking the residual offset that maximizes the likelihood of preference data $\mathcal{D}$ given prior $r^0$:
\begin{equation} \label{eq:map}
r'_{\psi} = \arg\max_{r'} \  \log p(\mathcal{D} \mid r^0 + r') + \log p(r^0 + r').
\end{equation}
Assuming a Gaussian prior on $r$ and Bradley-Terry likelihood \cite{bradley1952rank}, this corresponds to the optimization objective:
\begin{equation} \label{eq:obj}
\min_{r'_\psi} \ \mathcal{L}^r(\mathcal{D}, r^0 + r'_\psi) + \|r'_\psi\|^2,
\end{equation}
where $\mathcal{L}^r$ is the cross-entropy loss in \cref{eq:btloss}; in practice we replace the penalty $\|r'_\psi\|^2$ with a terminal \texttt{tanh} on $r'_\psi$. The prior $r^0$ serves as a baseline, while $r'_{\psi}$ learns an adaptive offset to align with human feedback. This structure enables both subtle refinements and large corrections, ensuring $\hat{r}^{\text{RRM}}_\psi = r^0 + r'_{\psi}$ effectively integrates prior knowledge.

\begin{algorithm}[tb]
\begin{small}
  \caption{Residual Reward Model Training}\label{alg:RRM}
  \begin{algorithmic}[1]
    \Require Reward batch size $M$, human feedback frequency $K$, \textcolor{blue}{pre-defined prior reward $r^0$}

    \State Initialize parameters of SAC ($Q_\theta, \pi_\phi$), \textcolor{blue}{reward $\hat{r}^{\text{RRM}}_\psi$}, buffer $\mathcal{D} \leftarrow \varnothing$
    \State \textcolor{blue}{// Unsupervised pre-training (optional, here we do not perform)}
    \For{\textit{each iteration}}
        \If{iteration $\%$ K == 0}
            \For{$m$ in $1,\dots, M$}
                \State Sample segment pair $(\sigma_0, \sigma_1) \sim \mathcal{B}$
                \State Query instructor for preference $y$
                \State Update preference buffer $\mathcal{D} \leftarrow \mathcal{D} \cup \{(\sigma_0,\sigma_1,y)\}$
            \EndFor
            \For{\textit{each gradient step}}
                \State Sample a minibatch $\{(\sigma_0,\sigma_1,y)\}_{j=1}^D \sim \mathcal{D}$
                \State Optimize $\mathcal{L}^r$ in \cref{eq:btloss} with respect to $\psi$
            \EndFor
            \State Relabel replay buffer $\mathcal{B}$ using \textcolor{blue}{$\hat{r}^{\text{RRM}}_\psi$ and $r^0$}
        \EndIf
        
        \For{\textit{each timestep} $t$}
            \State Collect $s_{t+1}$ by executing $a_t \sim \pi_\phi(a_t|s_t)$
            \State \textcolor{blue}{Compute current prior $r^0(s_t,a_t)$}
            \State \textcolor{blue}{Compute posterior reward (\cref{eq:rrm})}
            \State Store transition in $\mathcal{B}$:\\
            \hspace{0.5cm} $\mathcal{B} \leftarrow \mathcal{B} \cup \{(s_t, a_t, s_{t+1}, \textcolor{blue}{r^0, \hat{r}^{\text{RRM}}_\psi})\}$
        \EndFor
        \State Optimize SAC agent using data with reward $\hat{r}^{\text{RRM}}_\psi$ from $\mathcal{B}$
    \EndFor
  \end{algorithmic}
\end{small}
\end{algorithm}

\begin{figure}[t]
  \centering
  \includegraphics[width=\textwidth]{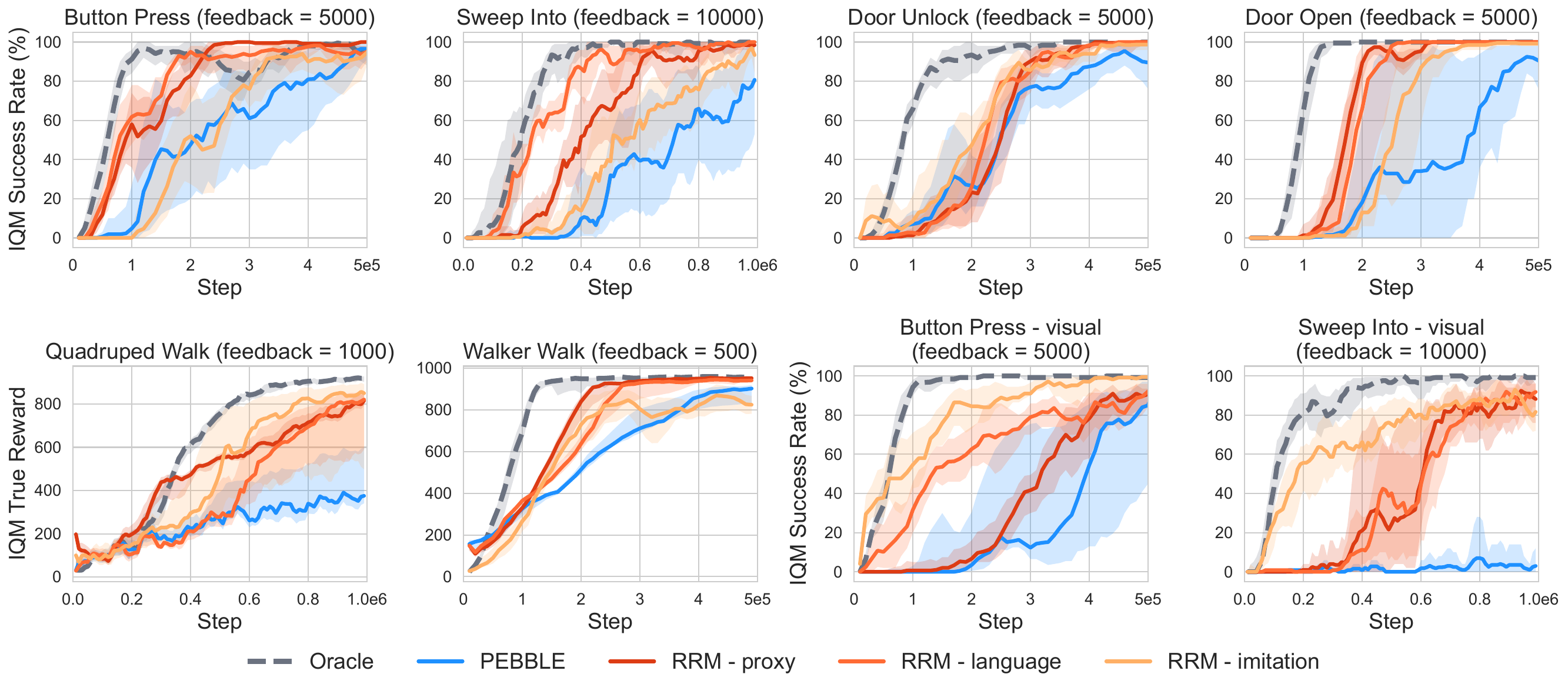}
  \caption{\textbf{Main Results.} We train RRMs and the baseline on 8 tasks across 10 seeds, using the same feedback schedule as Lee et al. \cite{lee2021pebblefeedbackefficientinteractivereinforcement}: 50 preferences every 5,000 steps for Meta-World, every 30,000 steps (20 times) for \textit{Quadruped Walk}, and every 20,000 steps (10 times) for \textit{Walker Walk}. The dotted blue line is the oracle; the shaded area is the interquartile range (IQR, $25$--$75\%$).}
  \label{fig:main}
\end{figure}

We instantiate $r'_{\psi}$ as an ensemble of three three-layer MLPs (256 hidden units each) built on PEBBLE's reward-model design \cite{lee2021pebblefeedbackefficientinteractivereinforcement} in our experiments, whose input concatenates the state--action pair with the prior value as $[s, a, r^0(s,a)]$, trained with learning rate $0.003$. \cref{fig:RRM} illustrates the high-level design of our method. The complete training procedure of our RRM, which integrates human preference queries with the Soft Actor-Critic (SAC) update loop, is summarized in \cref{alg:RRM} (\textcolor{blue}{blue text} highlights differences from PEBBLE \cite{lee2021pebblefeedbackefficientinteractivereinforcement}). Since the RRM framework is agnostic to prior origins, we investigate three practical sources for the prior reward $r^0$.

\subsection{Prior Rewards} \paragraph{Proxy reward (human-designed reward)} The proxy reward can be formulated by humans and it depends on the states of the environment \cite{laidlaw2024correlated}. Formally, it can be formulated as $r(s,a) = \sum_i w_i \phi_i(s,a)$, where $\phi_i(s,a)$ is the feature function and $w_i$ is constant. In state-based settings, we can define a proxy reward using information such as the target position, object position, and the agent's proprioceptive position. In contrast, we can only use the agent's proprioceptive position and the initial environmental setup data to obtain it in image-based settings.

\paragraph{Language reward (LLM-generated reward)} LLM is an important tool to generate reward for RL \cite{ma2024eureka}. The prompt $p$ is provided to LLM, including environmental information and task goal. Normally, LLM would generate a reward function $r(s,a)$ following $p$, which can be used as a prior reward.

\paragraph{Imitation reward} The prior reward can also be learned by imitation learning, especially inverse reinforcement learning (IRL) \cite{biyik2022demopref2}. Suppose we have demonstrations $d$, IRL can learn a reward model by fitting the similarity between the agent's behavior and the demonstrations. Once the reward model converges, we extract it as our prior reward.

\section{Experiments}
\label{sec:experiment}

We design our main experiments to answer these questions:

\begin{enumerate}[label=Q\arabic*:,leftmargin=10mm]
    \item Can RRMs improve the sample efficiency of existing PbRL algorithms? (\cref{sub:1}, \cref{sub:7})
    \item How sensitive is the performance to ablations of the RRM component rewards? (\cref{sub:2})
    \item How sensitive is the performance under strict real-world feedback constraints, such as sparse, noisy, or real human evaluators? (\cref{sub:3}, \cref{sub:8})
    % \item Can RRM successfully train the policy with preference label noise? (\cref{sub:4})
    % \item How can we choose an appropriate prior reward for RRM? (\cref{sub:5})
    \item Can RRM lead to higher success rates on a real robot faster than a baseline? (\cref{sub:6})
\end{enumerate}

% Now, we briefly summarize the results: our experiments demonstrate that RRMs improve the performance of several PbRL baselines by accelerating policy learning. Moreover, it effectively leverages diverse prior rewards, outperforming methods that rely only on prior reward learning, and remains robust even with priors with opposite or random semantics. Compared to PEBBLE, RRMs are less sensitive to infrequent and smaller amounts of feedback and remain stable under incorrect feedback. Finally, on a real manipulator, RRM reaches higher success rates in fewer steps than the baseline.

\subsection{Experimental Setup}

\begin{wrapfigure}{r}{0.55\textwidth}
  \centering
   \includegraphics[width=\linewidth]{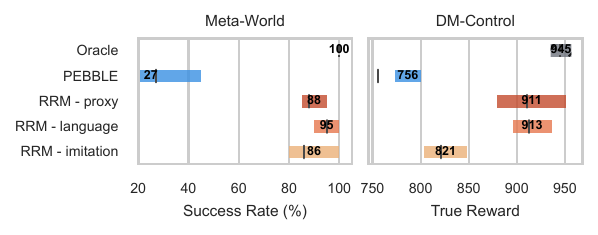}
    \caption{\textbf{IQM Performance in Meta-World and DM-Control.} In Meta-World, we choose the first step when a non-oracle method's 8/10 seeds reach 100$\%$ success rate for each task and compute the IQM success rate across these steps. In DM-Control, we directly utilize the final step for computing. The vertical line in the middle of the box represents the mean of the data, which is also labeled with a number. The edges of the box indicate the 25th and 75th percentiles. Oracle is illustrated by a dotted line.}
  \label{fig:mainbox}
\end{wrapfigure}

We choose four robotic manipulation tasks in Meta-World \cite{yu2020meta} and two locomotion tasks in DM-Control \cite{tunyasuvunakool2020} for our state-based evaluation: \textit{Button-press, Sweep-into, Door-open, Door-unlock, Quadruped-walk, Walker-walk}, together with two image-based manipulation tasks, yielding eight tasks in total. Similar to Lee et al.~\cite{lee2021pebblefeedbackefficientinteractivereinforcement}, the agent learns a reward model from a scripted teacher that provides a preference between two segments according to an unobserved ground-truth reward. For evaluation, we report the Inter-Quartile Mean (IQM) \cite{agarwal2021IQM} of the true success rate for Meta-World and the true accumulated return for DM-Control. IQM aggregates performance across the central 50\% of trials (the 25\%–75\% quantile range), making it more robust than simple averages to the high-variance and outlier runs common in preference-based RL.

%Consequently, we evaluate policy performance using the Inter-Quartile Mean (IQM) \cite{agarwal2021IQM} of the true success rate for Meta-World and the true accumulated return for DM-Control.

% We report the average IQM (Inter-Quartile Mean)  \cite{agarwal2021IQM} scores across different tasks, which refers to averaging the IQM scores at each evaluation step to assess the convergence speed and final IQM score of different methods. Furthermore, to intuitively evaluate the convergence performance of different methods, we select the first step at which the mean success rate over the top 75$\%$ of runs (25$\%$-100$\%$ quantile range) reaches its maximum for each non-oracle method, and present their IQM scores across every suite.

In our main state-based experiments, our method builds on PEBBLE \cite{lee2021pebblefeedbackefficientinteractivereinforcement} with an SAC \cite{haarnoja2018SAC} backbone. Prior rewards are normalized via $N(x)=\frac{x-\mu}{\sigma+\epsilon}$ ($\mu, \sigma$: training-time empirical statistics) to align their magnitude with the residual network $r'_{\psi}$. Since $r'_{\psi}$ is bounded to $[-1, 1]$ via a terminal tanh activation, this normalization is applied consistently across all prior reward types (proxy, language, and imitation) to ensure optimization stability. The three prior types are constructed as follows: (1) \textbf{Proxy rewards} encode task heuristics: negative distances from the end-effector of the robotic arm to the object and distance from the position of the object to the goal for Meta-World (${r_p =N (-\Vert s_{\text{obj}} - g\Vert - \Vert s_{\text{ee}} - s_{\text{obj}}\Vert)}$), and an upright torso penalty for DM-Control (${r_p =N(x)}$). (2) \textbf{Language rewards} are code-executable, piecewise functions generated by GPT-5 \cite{gpt5} from the task description and the environment's state–action definitions; among 5 candidates, we keep the best-executing one and freeze it as the prior $r^0$ \cite{ma2024eureka}. (3) \textbf{Imitation rewards} are extracted from Adversarial Inverse Reinforcement Learning (AIRL) \cite{fu2017airl} models pre-trained on 50 demonstrations.

Image-based RRM is implemented on DrQv2 \cite{yarats2021DrQv2} and DrM \cite{xu2023drm}. Lacking explicit object states, proxy and language priors rely strictly on robot proprioception and initial workspace setups. The proxy reward penalizes actions outside a feasible bounding region: $r_p = -1$, if $s_{\text{ee}}\notin \mathcal{F}$. Specifically, this region $\mathcal{F}$ is the union of the rectangular area formed from the end-effector to the object’s initial position $\mathcal{F}_1$ and the area from the object’s initial position to the target $\mathcal{F}_2$. GPT-5 language reward can be roughly written in the form of $r_l = N(-\Vert s_{\text{ee}}-s_{\text{obj-init}}\Vert - \Vert s_{\text{ee}}-g\Vert)$. For image imitation rewards, we train AIRL on 100 demonstrations; to ensure that the pre-trained reward model is used effectively, its coupled representation encoder is transferred directly to the policy learning phase.

\subsection{Main Results}
\label{sub:1}

\begin{figure}[t]
    \centering
    \begin{minipage}{0.49\textwidth}
        \centering
        \includegraphics[width=\textwidth]{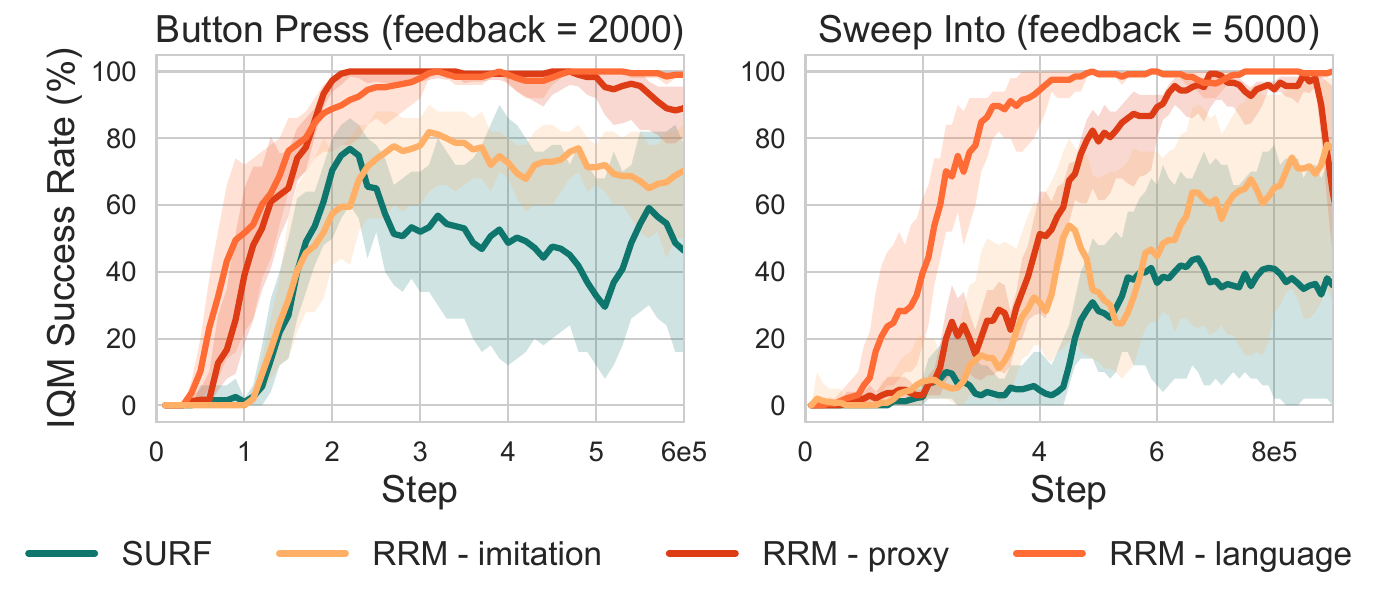}
        \subcaption{Experiments on SURF.}
        \label{fig:base-a}
    \end{minipage}
    \begin{minipage}{0.5\textwidth}
        \centering
        \includegraphics[width=\textwidth]{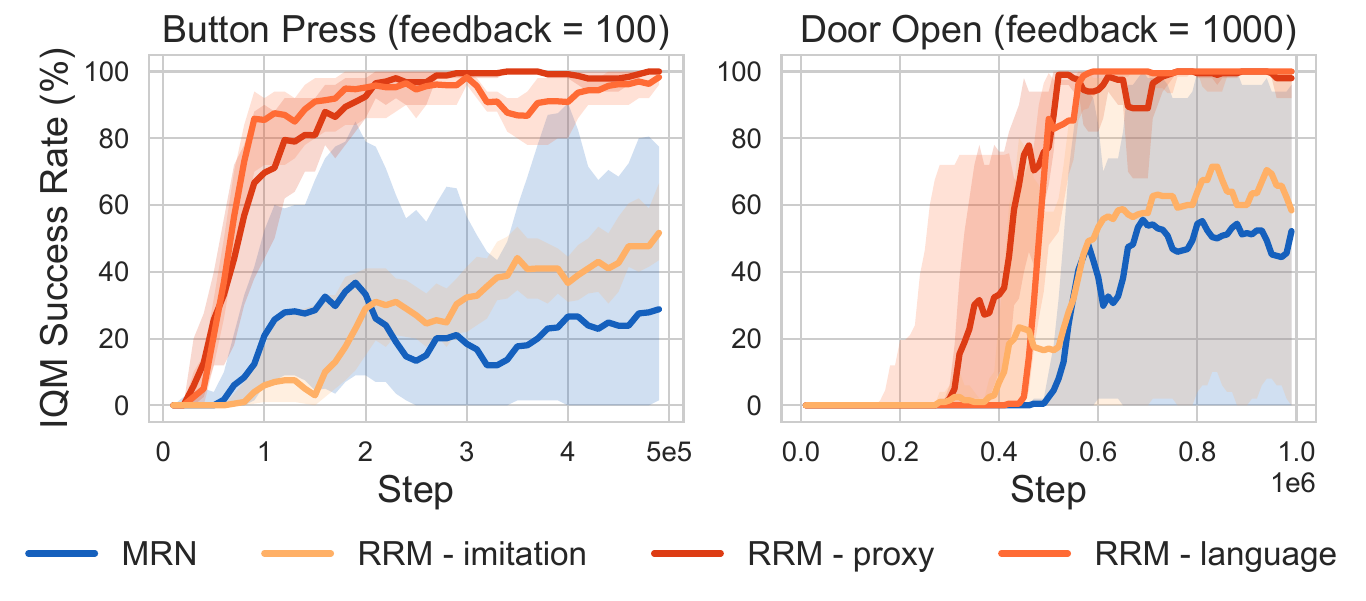}
        \subcaption{Experiments on MRN.}
        \label{fig:base-b}
    \end{minipage}
    \caption{\textbf{Applying RRM to Other PbRL Methods.} The shaded area indicates the interquartile range (IQR). (a) Comparison between RRMs based on SURF and SURF \cite{park2022surf}. (b) Comparison between RRMs based on MRN and MRN \cite{liu2022mrn}.}
    \label{fig:base}
\end{figure}

We compare RRMs with their baseline PEBBLE and SAC with true reward (Oracle) using IQM success rates and true episode returns on different tasks, illustrated in \cref{fig:main}. Furthermore, we draw an intuitive box plot in \cref{fig:mainbox} by selecting the first step when a non-oracle method reaches its maximum in Meta-World and the final step in DM-Control. In Meta-World, RRM with language reward achieves the highest success rate and fastest policy convergence, confirming LLMs' capability to generate well-aligned rewards. RRM with proxy reward also achieves strong performance since heuristic guesses usually effectively guide correct policy learning in such manipulation tasks. Conversely, RRM with imitation reward is slightly worse. This stems from IRL demonstration bias: incomplete demonstrations or distribution shift restrict the policy to demonstrated trajectories, limiting exploration and making residual learning more difficult. Nevertheless, all RRM variants significantly outperform PEBBLE, demonstrating their effectiveness.

Results in DM-Control locomotion tasks are similar to the manipulation tasks. However, in \textit{Quadruped Walk}, RRM with imitation reward achieves a higher return but larger variance, likely due to the task's kinetic sensitivity, which makes reliance on demonstrations highly sensitive to initialization. Additionally, RRM with language reward converges more slowly than the proxy, showing LLMs struggle in constructing locomotion dynamics.

In visual tasks, RRM with imitation reward performs the best, even close to the Oracle—likely benefiting from the visual encoder transfer. Apart from that, RRM with proxy reward and language reward outperforms PEBBLE; the language reward is more capable of capturing reward information from initialized positions than the penalty proxy. Overall, while all prior types improve sample efficiency, their variance shows that choosing an appropriate prior is crucial.

\begin{figure}[t]
    \centering
    \begin{minipage}{0.515\textwidth}
        \centering
        \includegraphics[width=\textwidth]{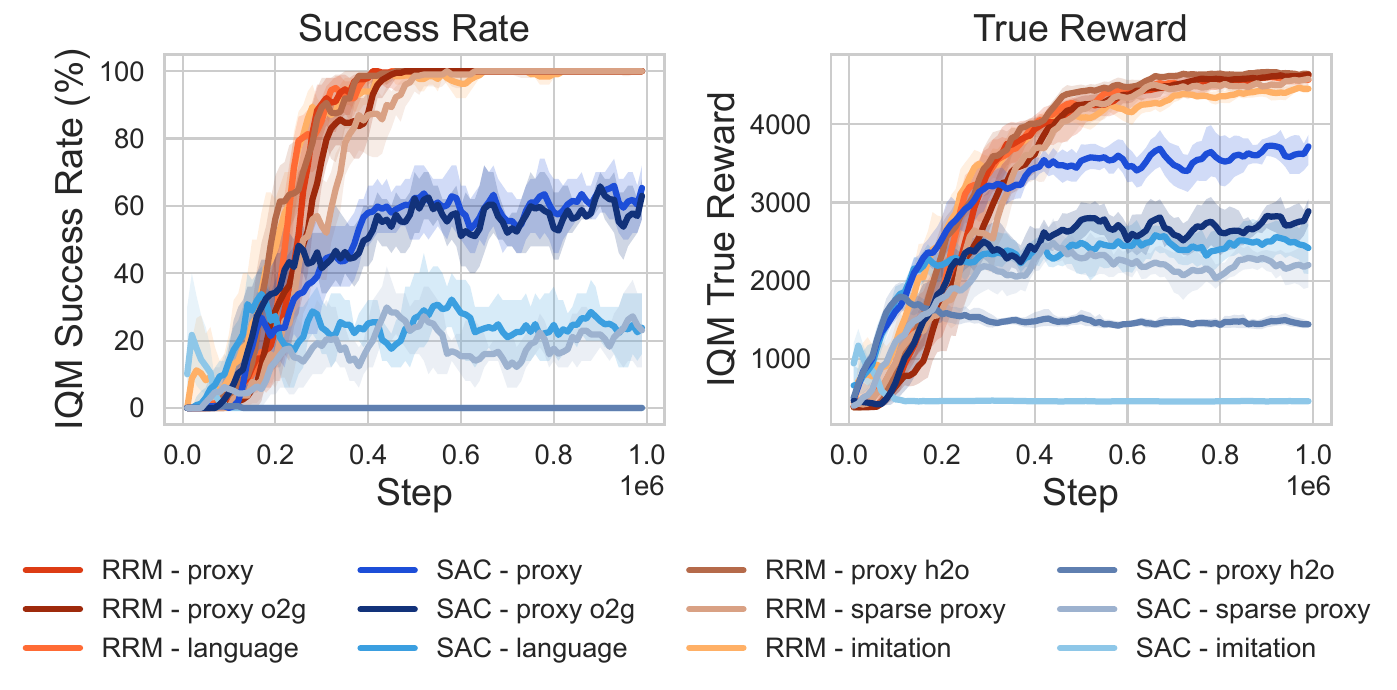}
        \subcaption{Ablation on residuals.}
        \label{fig:abla-a}
    \end{minipage}
    \begin{minipage}{0.475\textwidth}
        \centering
        \includegraphics[width=\textwidth]{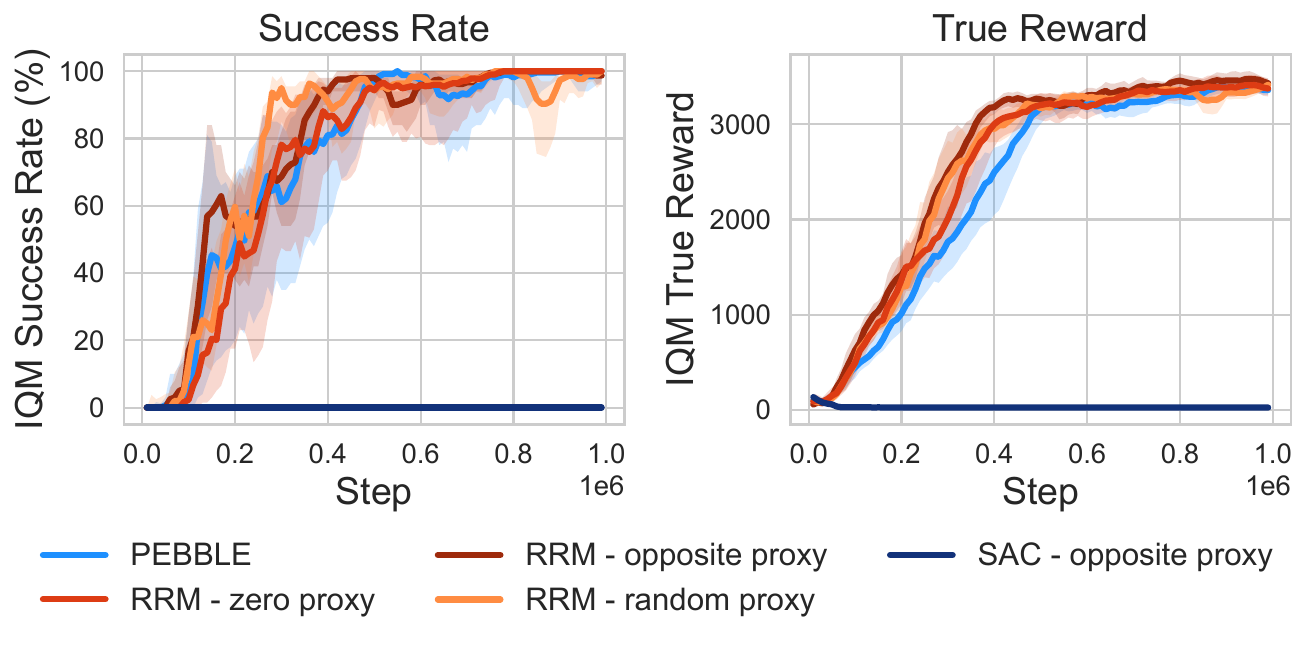}
        \subcaption{Ablation on priors.}
        \label{fig:abla-b}
    \end{minipage}
    \caption{\textbf{Ablation Studies.} The shaded area indicates the interquartile range (IQR). (a) Contribution of residual rewards in RRM with different prior rewards. Proxy o2g means the normalized negative distance from the position of the object to the target in the task and proxy h2o indicates the normalized negative distance from the end-effector of the robotic arm to the object. (b) Robustness of RRM to unexpected prior rewards. Random proxy indicates a random value with the same scale as the residual reward.}
    \label{fig:abla}
\end{figure}

\subsection{Applying RRM to Other PbRL Methods}
\label{sub:7}

To demonstrate its orthogonality, we integrate RRM into SURF \cite{park2022surf} and MRN \cite{liu2022mrn}. Following Park et al.~\cite{park2022surf} and Liu et al.~\cite{liu2022mrn}, we adopt the same feedback budgets for our experiments: 2,000 and 5,000 queries for \textit{Button-press} and \textit{Sweep-into} (SURF), and 1,000 and 100 queries for \textit{Door-open} and \textit{Button-press} (MRN), respectively. As illustrated in \cref{fig:base}, RRMs can consistently improve the base performance of PbRL algorithms, confirming their broad applicability as a plug-and-play module for preference-based reward updating methods.

\begin{figure*}[t]
  \centering
  \begin{minipage}[htbp]{0.52\textwidth}
    \centering
    \captionof{table}{\textbf{Success Rate ($\%$) in Stochastic and Mistaken Settings.} We use PEBBLE in the stochastic setting trained for 300,000 steps and MRN in the mistaken setting trained for 500,000 steps. The table reports the final success rate $\pm95\%$ confidence interval after training for a fixed number of steps on \textit{Button-press}.}
    \label{tab:wrongiqm1}
    \resizebox{\linewidth}{!}{
    \begin{tabular}{lccc}
    \toprule
    {\textbf{Method}} & \textbf{Stochastic} & \textbf{10$\%$ Mistaken} & \textbf{15$\%$ Mistaken}\\ 
    \midrule
    Oracle & 99.3\scriptsize{$\pm$4.1} & 100.0\scriptsize{$\pm$1.3} & 100.0\scriptsize{$\pm$1.3} \\
    \midrule
    PbRL baselines & 71.8\scriptsize{$\pm$5.3} & 16.0\scriptsize{$\pm$1.7} & 2.3\scriptsize{$\pm$0.3}\\ 
    RRM - proxy & \textbf{98.5\scriptsize{$\pm$5.9}}  & \textbf{98.8\scriptsize{$\pm$10.4}} & 46.0\scriptsize{$\pm$4.4}\\
    RRM - language & 94.9\scriptsize{$\pm$5.4} & 91.3\scriptsize{$\pm$8.1} & \textbf{48.7\scriptsize{$\pm$4.1}} \\
    RRM - imitation & 86.6\scriptsize{$\pm$6.4} & 46.5\scriptsize{$\pm$5.9}  & 28.8\scriptsize{$\pm$3.4}\\
    \bottomrule
    \end{tabular}
    }
  \end{minipage}
  \hfill
  \begin{minipage}[htbp]{0.45\textwidth}
    \centering
    \includegraphics[width=\linewidth]{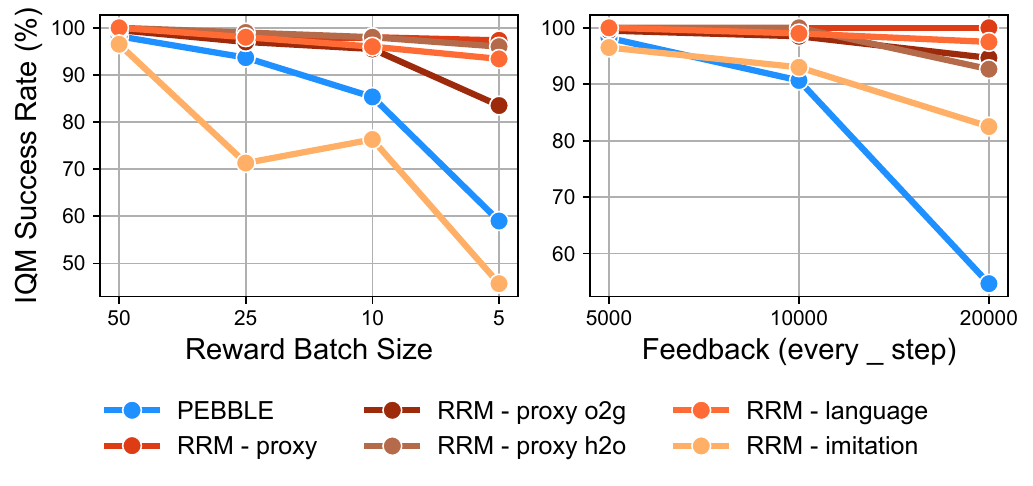}
    \caption{\textbf{Results with Less Feedback.} The curves represent the mean of the IQM success rate of algorithms with limited feedback.}
    \label{fig:less}
  \end{minipage}
\end{figure*}
\subsection{Ablation Studies}
\label{sub:2}

To evaluate the contribution of each RRM component, we conduct two sets of ablation experiments on \textit{Door-unlock} and \textit{Button-press}. In the first set, we ablate the residual reward by training SAC solely on the prior rewards. As shown in \cref{fig:abla-a}, these prior-only baselines achieve a maximum success rate of 60$\%$, whereas the full RRM leverages even incomplete (e.g., hand-to-object) or sparse proxy rewards to reach $100\%$, significantly outperforming PEBBLE and prior-only SAC. In the second set, we ablate the prior reward. As depicted in \cref{fig:abla-b}, RRM with zero proxy reward initially trails PEBBLE because RRM removes PEBBLE's unsupervised pre-training phase and only relies on the prior reward for initial guidance. Interestingly, when tested with an opposite proxy reward, i.e., a negated version of the proxy reward, despite SAC not being able to learn from it, RRM still achieves a $100\%$ success rate. It is because RRM learns to correlate lower proxy values with preferred behaviors, optimizing the residual model to effectively invert the negated prior. Furthermore, RRM maintains strong robustness even when initialized with purely random rewards.

\subsection{Robustness Analysis}
\label{sub:3}

To evaluate RRM's robustness to sparse data, we restrict the feedback budget provided by the scripted teacher. The original Meta-World setup updates the reward model with 50 preferences every 5,000 steps (accumulating 5,000 queries over 500,000 steps). We reduce this by either decreasing the query batch size to 25, 10, or 5 or lowering the query frequency to every 10,000 or 20,000 steps. As shown in \cref{fig:less}, PEBBLE's success rate degrades significantly, dropping below 60\% when the reward batch size is reduced to 5 or feedback is provided every 20,000 steps. In contrast, RRM demonstrates greater stability, particularly with the proxy and language priors, all of which consistently achieve at least an 85\% success rate despite reduced feedback. However, RRM with imitation reward struggles under small reward batches, as its limited state-space coverage gives insufficient guidance in unexplored regions, increasing reliance on the residual reward to refine the policy through feedback.

Unlike preferences from the true reward, real humans often provide noisy or irrational feedback, and standard PbRL is vulnerable to such stochastic and mistaken preferences \cite{lee2021bpref}. We therefore evaluate RRMs on \textit{Button-press} under these two noise models. Results shown in \cref{tab:wrongiqm1} indicate RRMs maintain robust performance under stochastic noise and 10\% mistaken feedback. While 15\% mistaken feedback degrades performance, RRMs still demonstrate clear improvements over baselines. This suggests that RRMs rely on accurate preference labels to correct the prior so that severe human errors inevitably affect performance. Nevertheless, RRMs generally maintain higher resilience against abnormal feedback than standard PbRL due to the prior reward.

\begin{figure}[t]
\centering
    \begin{minipage}[b]{0.29\textwidth}
        \centering
        \includegraphics[width=\textwidth]{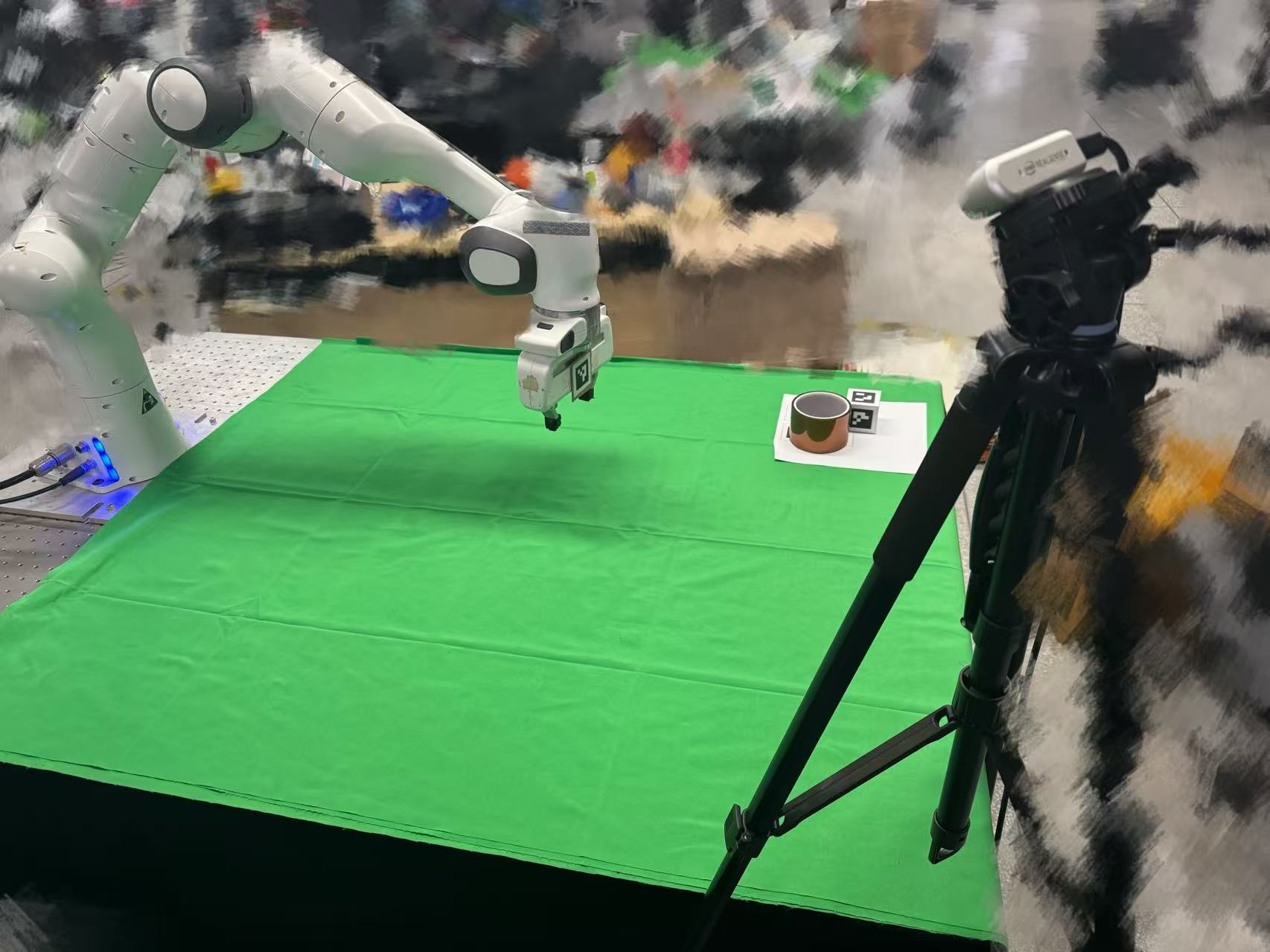}
        \subcaption{Real robot setup.}
        \label{fig:real-a}
    \end{minipage}
    \begin{minipage}[b]{0.7\textwidth}
        \centering
        \includegraphics[width=\textwidth]{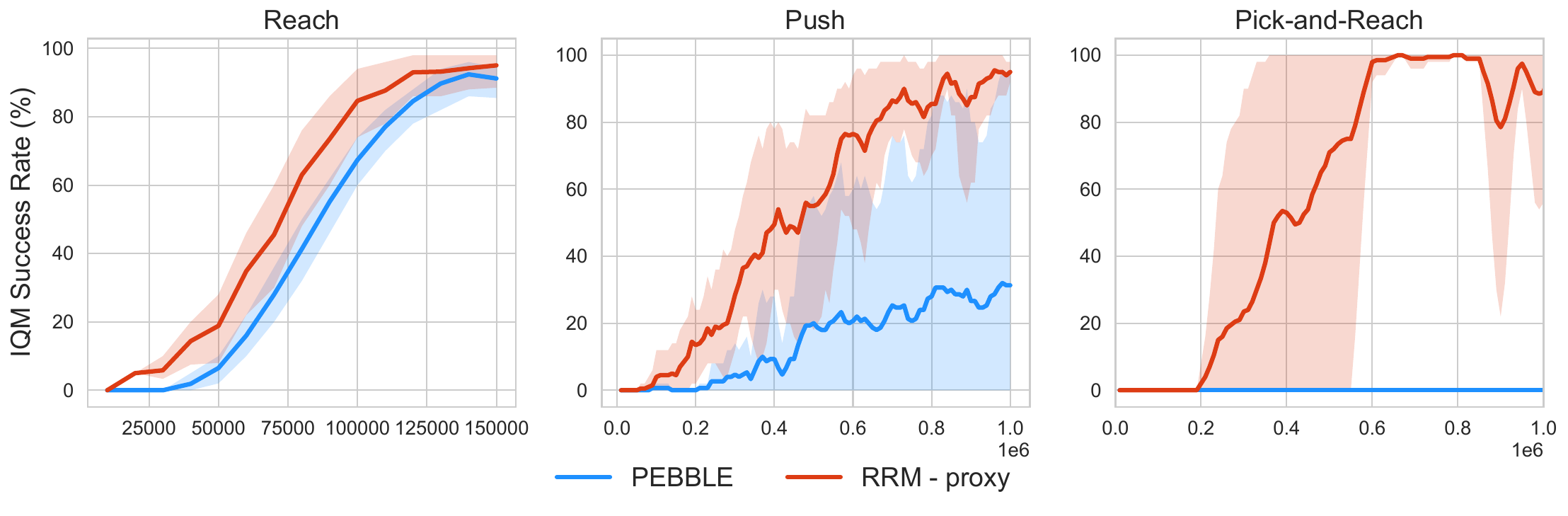}
        \subcaption{Simulation learning curves used for real-world deployment.}
        \label{fig:real-b}
    \end{minipage}
  \caption{\textbf{Real World Setup.} (a) We use a Franka Emika Panda arm, Franka Hand gripper and two Intel D435 RealSense cameras in a multi-view setting. We calibrate the cameras and obtain the block's pose using ArUco markers. (b) We train the policy on the simulation built by PyBullet for sim-to-real. Reach, Push and Pick-and-Reach are trained for 150,000 and 1,000,000 steps. 50 preferences are provided every 5,000 steps.}
  \label{fig:real}
\end{figure}

\subsection{Evaluation with Real Human Feedback}
\label{sub:8}

\begin{wrapfigure}[16]{r}{0.32\textwidth}
  \centering
    \includegraphics[width=\linewidth]{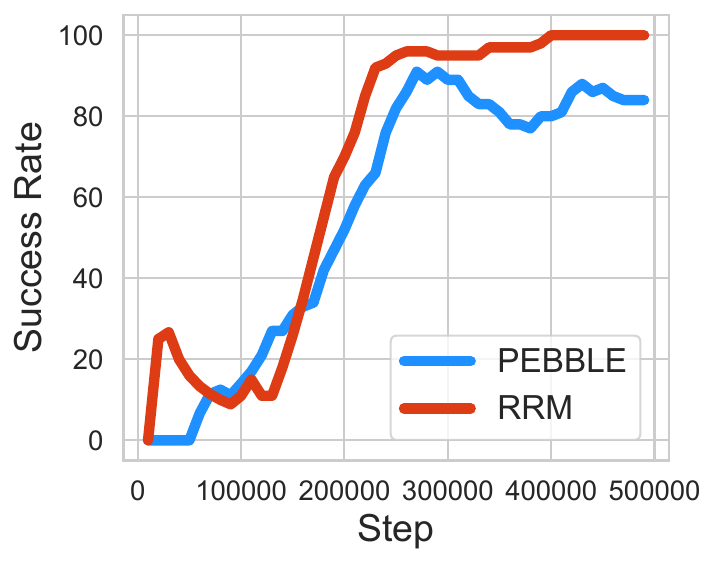}
    \caption{\textbf{Real Human Preference Results.} RRM - proxy h2o is compared against PEBBLE on the \textit{Window-open} task using strictly limited real human feedback. Curves represent the smoothed success rate across 5 seeds.}
    \label{fig:realhumanexp}
\end{wrapfigure}

To further validate RRM's practical applicability, we conduct a human-in-the-loop experiment on \textit{Window-open}. We build a lightweight interface that renders each queried segment pair as two side-by-side animations, with the cumulative reward of each segment shown on top only as a soft reference, since humans do not judge preferences by return alone. Five volunteers label the queries by pressing the left or right arrow key to select the preferred segment, and are asked to avoid ties so that every query yields a decision. Under strict real-world deployment constraints, human evaluators provide a sparse budget of 400 total preference queries, i.e., 20 queries every 10,000 steps during the first 200,000 steps. As \cref{fig:realhumanexp} shows, while PEBBLE struggles under such minimal supervision, RRM leverages its heuristic prior reward to achieve faster convergence and higher success rates. It demonstrates robust feedback efficiency even under limited, noisy, and irregular human supervision.

\subsection{Real World Experiments}
\label{sub:6}

We further demonstrate RRM's practical efficacy via zero-shot sim-to-real transfer on a physical Franka Emika Panda manipulator across three tasks: Reach, Push, and Pick-and-Reach. We train our policies in PyBullet environments that mirror our physical setup (\cref{fig:real-a}) \cite{coumans2021, ablett2024vpace}. The action space consists of continuous end-effector Cartesian velocities (3D for Reach and Pick-and-Reach; 2D for Push), while the observation space includes the end-effector position, object pose (including yaw for Push and Pick-and-Reach), and target goal coordinates. To obtain accurate object pose coordinates, we attach an ArUco marker to the cube and use an external camera system for state estimation during real-world deployments. Using identical hyperparameters to our Meta-World experiments, we generate synthetic preference labels during simulation from the true task rewards: $r_{\text{reach}} = -\Vert s_{\text{ee}} - s_{\text{obj}}\Vert$ and $r_{\text{push/pick}} = 3\times(1 - \tanh(10\times \Vert s_{\text{goal}} - s_{\text{obj}}\Vert)) + (1 - \tanh(10\times \Vert s_{\text{obj}} - s_{\text{ee}}\Vert))$. For RRM's prior rewards, we employ a minimal 1D proxy penalizing y-axis distance for Reach ($r_{\text{proxy}} = -\Vert s_{\text{ee},y} - s_{\text{goal}, y}\Vert$), and a heuristic hand-to-object distance prior ($r_{\text{proxy}} = - \Vert s_{\text{obj}} - s_{\text{ee}}\Vert$) for Push and Pick-and-Reach.

Under this configuration, we train the Reach task for 150,000 steps, the Push and Pick-and-Reach tasks for 1,000,000 steps before direct transfer to the real world. As detailed in \cref{fig:real-b}, RRM consistently outperforms PEBBLE across all three tasks. In the complex Push and Pick-and-Reach tasks, RRM achieves 70\% and 85\% physical success rates at 600,000 simulation steps, respectively, demonstrating significantly faster policy convergence; per-checkpoint real-robot numbers are reported in \cref{tab:realiqm}. In contrast, PEBBLE either fails to make progress or achieves a success rate of only 50\% under the same training constraints. These experiments confirm that RRM significantly accelerates policy convergence and improves reward learning for real-world applications.

\begin{table}[t]
\centering
\caption{\textbf{Real-robot success rate ($\%$) and return.} Policies trained in simulation are transferred zero-shot to the physical Franka Panda at several simulation-training checkpoints; \cref{fig:real-b} shows the corresponding simulation learning curves. Each entry averages 5 seeds $\times$ 20 real-world episodes. Reach and the two harder tasks are evaluated at different checkpoints, matching their different training budgets. Prior names follow \cref{tab:priors}.}
\label{tab:realiqm}
\small
\setlength{\tabcolsep}{3pt}
\renewcommand{\arraystretch}{1.15}
\begin{tabular*}{\textwidth}{@{\extracolsep{\fill}} l l cc cc cc @{}}
\toprule
 & & \multicolumn{6}{c}{\textbf{Simulation training steps}} \\
\cmidrule(l){3-8}
\multirow{2}{*}{\textbf{Task}} & \multirow{2}{*}{\textbf{Method}}
  & \multicolumn{2}{c}{\textbf{50{,}000}} & \multicolumn{2}{c}{\textbf{100{,}000}} & \multicolumn{2}{c}{\textbf{150{,}000}} \\
\cmidrule(lr){3-4} \cmidrule(lr){5-6} \cmidrule(lr){7-8}
 & & Success\,$\uparrow$ & $R\uparrow$ & Success\,$\uparrow$ & $R\uparrow$ & Success\,$\uparrow$ & $R\uparrow$ \\
\midrule
\multirow{2}{*}{Reach} & PEBBLE           & 33.3 & -10.9 & \textbf{93.3} & -5.7 & 95.0 & -5.0 \\
                       & RRM -- 1D proxy  & \textbf{45.0} & \textbf{-7.3} & \textbf{93.3} & \textbf{-4.3} & \textbf{100.0} & \textbf{-3.3} \\
\midrule
\addlinespace[1pt]
 & & \multicolumn{2}{c}{\textbf{400{,}000}} & \multicolumn{2}{c}{\textbf{600{,}000}} & \multicolumn{2}{c}{\textbf{800{,}000}} \\
\cmidrule(lr){3-4} \cmidrule(lr){5-6} \cmidrule(lr){7-8}
 & & Success\,$\uparrow$ & $R\uparrow$ & Success\,$\uparrow$ & $R\uparrow$ & Success\,$\uparrow$ & $R\uparrow$ \\
\midrule
\multirow{2}{*}{Push} & PEBBLE              & 40.0 & -22.7 & 50.0 & -19.0 & 90.0 & -17.3 \\
                      & RRM -- proxy h2o    & \textbf{41.6} & \textbf{-16.8} & \textbf{70.0} & \textbf{-16.8} & \textbf{95.0} & \textbf{-15.4} \\
\addlinespace[3pt]
\multirow{2}{*}{Pick-and-Reach} & PEBBLE            & 0.0 & -35.8 & 0.0 & -32.2 & 0.0 & -31.2 \\
                                & RRM -- proxy h2o  & 0.0 & \textbf{-30.9} & \textbf{85.0} & \textbf{-24.5} & \textbf{95.0} & \textbf{-23.5} \\
\bottomrule
\end{tabular*}
\end{table}
\section{Conclusion}
\label{sec:conclusion}

This paper introduces the Residual Reward Model (RRM) to accelerate policy convergence in Preference-based Reinforcement Learning (PbRL) for robot control. By structurally integrating domain prior knowledge with residual human feedback, RRM achieves substantial sample efficiency gains in enhancing PbRL algorithms across state- and image-based benchmarks. Crucially for physical deployments, RRM exhibits robustness against defective prior rewards and insufficient human feedback. Finally, our sim-to-real experiments confirm that RRM significantly accelerates real-world policy convergence on a physical manipulator. In general, RRM offers a highly practical, plug-and-play solution to overcome the feedback bottleneck in preference-based robot learning.

% Despite these advantages, how to obtain high-quality prior rewards remains an open problem, and the reliance on human feedback continues to limit large-scale deployment. Developing methods that can automatically generate informative feedback or construct reliable priors remains an important direction for future work.

\bibliographystyle{plainnat}
\bibliography{rrm}

\newpage

\beginappendix
%%%%%%%%%%%%%%%%%%%%%%%%%%%%%%%%%%%%%%%%%%%%%%%%%%%%%%%%%%%%%%%%%%%%%%%%%%%%%%%%
\section{Method Details}
\label{app:method}

\subsection{Motivating RRMs with MAP estimation}
\label{app:bayes_formulation}

Our goal is to estimate a reward function from data using a reward prior:
\begin{align}
p(r | \mathcal D) \propto p(\mathcal D | r) \cdot p(r), \label{eq:bayes}
\end{align}
where $p(r)$ is an arbitrary reward prior, $p(\mathcal D | r)$ is the likelihood of observing the preference data $\mathcal D = \{(\sigma_0^i, \sigma_1^i, y^i)\}_{i=1}^N$ given a reward function $r$, and $p(r | \mathcal D)$ is the posterior over reward functions that we seek to estimate. By assuming independence over each example and using the reward estimator $\hat{r}_\psi$ with the Bradley-Terry model (\cref{eq:btmodel}), $p(\mathcal D | r) = \prod_{i=1}^N P_\psi(\sigma_0^i \succ \sigma_1^i)^{y(0)} \cdot  P_\psi(\sigma_1^i \succ \sigma_0^i)^{y(1)}$. Let $p(r) = \mathcal N(r; r^0, \sigma^2)$. Recall the definition of the residual reward model (\cref{eq:rrm}): $\hat{r}_\psi^{\text{RRM}} = r^0 + r'_{\psi}$, where the residual $r'_{\psi}$ is learned, $r^0$ is given, and the combined model to be estimated is $\hat{r}_\psi^{\text{RRM}}$. Now, performing MAP estimation of \cref{eq:bayes}:
\begin{align}
\text{argmax}_{r} &p(r| \mathcal D) \\ 
&=\text{argmax}_r \log p(\mathcal D | r) + \log p(r) \nonumber\\ 
&= \text{argmax}_r \log p(\mathcal D | r) - \frac{1}{2\sigma^2} (r - r^0 )^2 \nonumber\\ 
&=  \text{argmax}_r \sum_{i=1}^N \left(y(0) \cdot \log P_\psi(\sigma_0^i \succ \sigma_1^i)+ y(1) \cdot \log P_\psi(\sigma_1^i \succ \sigma_0^i)\right) - \frac{1}{2\sigma^2}(r'_\psi)^2\nonumber \\
&= \text{argmax}_r \frac{1}{N}\sum_{i=1}^N \left(y(0) \cdot \log P_\psi(\sigma_0^i \succ \sigma_1^i)+ y(1) \cdot \log P_\psi(\sigma_1^i \succ \sigma_0^i)\right)- \frac{1}{2\sigma^2}(r'_\psi)^2 \nonumber \\
&\approx \text{argmin}_r \mathcal L^r + \frac{1}{2\sigma^2}(r'_\psi)^2, \label{eq:map_estimate}
\end{align}

where the approximation results from a finite-sample estimate of the expectation in $\mathcal L^r$. In practice, we use \texttt{tanh} to enforce small values of $r'_\psi$ (see \cref{app:model_structure}), instead of the squared penalty term.

\subsection{A toy example for the prior and the residual reward}
\label{app:toy}

In a car planning and obstacle avoidance task, the task description is: ``The car needs to drive around obstacles and reach the goal location.'' From this description, we can quickly infer: 1. The car must avoid obstacles, and 2. The car must reach the goal area. So the reward function that is fitted based on these two objectives is the prior reward. If we use the simplest sparse reward function to represent these two pieces of information, the prior reward could be written as:
\begin{equation}
    r^0(s,a) = \begin{cases}-1 & \text { if } \Vert s-s_o \Vert < 0.1, \forall s_o\in \mathcal{O}, \\ 10 & \text { if } \Vert s-g \Vert < 0.5, \\ 0 & \text { otherwise },\end{cases}
\end{equation}
where $\mathcal{O}$ denotes obstacle states and $g$ is the goal. 

However, a prior reward alone often cannot encompass all the necessary information. In this example, our ultimate goal is to have the car avoid obstacles and reach the target point as quickly as possible. Nonetheless, achieving this determination by any means is not what we want, and could even make it difficult to find the optimal solution to the problem. Due to factors such as the car's engine, movement mode, the shape of obstacles, and the relative position between the location of the target point and the obstacle, we want the car to avoid obstacles at a safe speed and choose a path that minimizes speed loss while efficiently reaching the target area. However, these choices are difficult to specify directly. This information always needs to be learned through exploration by the agent itself. Therefore, the task goal information implied by the combined effect of these environmental factors is the learned reward $r'(s,a)$ for the task.

\subsection{Image-based RRM}
\label{app:imagerrm}

\begin{figure}[htbp]
  \centering
  \includegraphics[width=\textwidth]{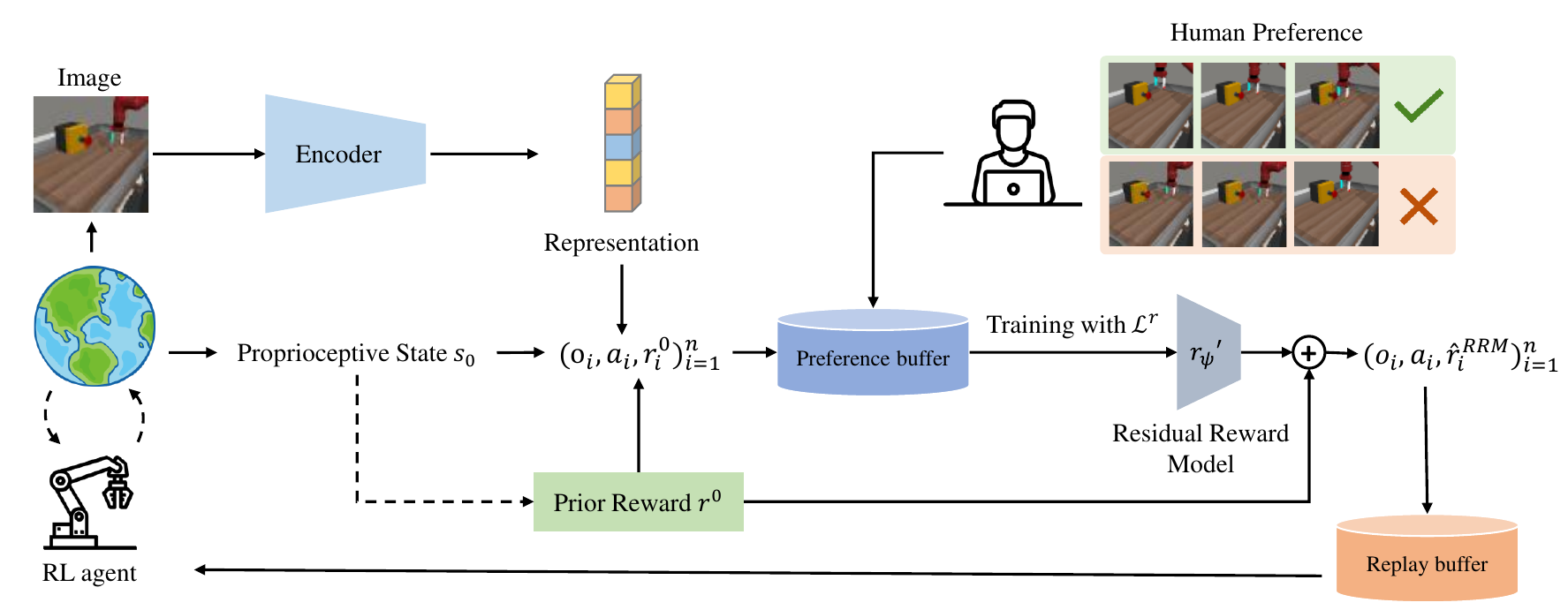}
  \caption{\textbf{Image-based Residual Reward Model.} In image-based settings, RRM obtains images and proprioceptive states from the environment rather than states. An encoder is used for extracting representations from images and is  jointly trained with the RL agent.}
  \label{fig:RRM_i}
\end{figure}

We depict image-based RRM in \cref{fig:RRM_i}. Compare to the state-based RRM in \cref{fig:RRM}, it adds an encoder to get the representations of images. Meanwhile, the proprioceptive state is used to provide information to the prior reward, enabling vision-based agent learning.

The corresponding training procedure is given in \cref{alg:RRMV}, which differs from \cref{alg:RRM} only in the encoder $f_\xi$ and in whether the prior is evaluated on the proprioceptive state or on the learned representation.

\begin{algorithm}[H]
\begin{small}
  \caption{Image-based Residual Reward Model}\label{alg:RRMV}
  \begin{algorithmic}[1]
    \Require reward batch size $M$, frequency of human feedback $K$, \textcolor{blue}{a pre-defined prior reward $r^0$, proprioceptive state $s$, image observation $o$}

    \State Initialize parameters of $Q_\theta$, encoder $f_\xi$, $\hat{r}_\psi$ and preference buffer $\mathcal{D} \leftarrow \varnothing$
    \State \textcolor{blue}{// Unsupervised pertaining (optional)}
    \For{\textit{each iteration}}
        \If{iteration $\%$ K == 0}
            \For{$m$ in 1,$\dots, M$}
                \State $(\sigma^0, \sigma^1) \sim \mathcal{D}$
                \State Query instructor for $y$
                \State Store preference $\mathcal{D} \leftarrow \mathcal{D} \cup \{(\sigma^0,\sigma^1,y)\}$
            \EndFor
            \For{\textit{each gradient step}}
                \State Sample a minibatch $\{(\sigma^0,\sigma^1,y)\}_{j=1}^D \sim \mathcal{D}$
                \State Optimize $L^r$ in \cref{eq:btloss} with respect to $\psi$
            \EndFor
            \State Relabel entire replay buffer $\mathcal{B}$ using $\hat{r}_\psi$ \textcolor{blue}{and $r^0(s_t,a_t)$ or $r^0(f_\xi(o_t),a_t)$}
        \EndIf
        \For{\textit{each timestep} $t$}
            \State Collect $o_{t+1},s_{t+1}$ by taking $a_t \sim \pi_\phi(a_t|o_t)$
            \State \textcolor{blue}{Compute current prior reward $r^0(s_t,a_t)$ or $r^0(f_\xi(o_t),a_t)$}
            \State \textcolor{blue}{Compute posterior reward $\hat{r}^{\text{RRM}}(s_t,o_t,a_t) = r^0(s_t,a_t) + \hat{r}_\psi(f_\xi(o_t),a_t,r^0(s_t,a_t))$ or $\hat{r}^{\text{RRM}}(o_t,a_t) = r^0(f_\xi(o_t),a_t) + \hat{r}_\psi(f_\xi(o_t),a_t,r^0(f_\xi(o_t),a_t))$}
            \State Store transitions $\mathcal{B} \leftarrow \{(o_t,s_t,a_t,o_{t+1},s_{t+1},\textcolor{blue}{r^0(s_t,a_t),\hat{r}^{\text{RRM}}(s_t,o_t,a_t)}\}$ or $\mathcal{B} \leftarrow \{(o_t,s_t,a_t,o_{t+1},s_{t+1},\textcolor{blue}{r^0(f_\xi(o_t),a_t),\hat{r}^{\text{RRM}}(o_t,a_t)}\}$
        \EndFor
        \State Optimize DrQ-v2 agent using $\mathcal{B}$
    \EndFor
    
  \end{algorithmic}
\end{small}
\end{algorithm}

\subsection{Model structure}
\label{app:model_structure}

We build our residual reward network upon the PEBBLE reward model structure \cite{lee2021pebblefeedbackefficientinteractivereinforcement}. The residual reward network is an ensemble of three reward networks, using three layers with 256 hidden units. At the end of each network, we use a tanh function to bound the output within $(-1,1)$. The networks are trained by cross-entropy loss defined by \cref{eq:btloss} with learning rate $3\times10^{-4}$ in the state-based setting and $1\times10^{-4}$ in the image-based setting.

In image-based RRM, we follow DrQ-v2 \cite{yarats2021DrQv2} to build the encoder and the agent. The encoder applies four convolutional layers, each followed by a ReLU activation, to extract feature maps from the image. The first convolutional layer has a $3\times3$ kernel size, 2 strides with 32 output dimensions, and the rest have the same kernel size, with 1 stride with 32 input and output dimensions. The output is flattened into a 1D vector, which is 20000 dimensions in our experiments. Meanwhile, there are fully connected layers in front of the actor and critic to further reduce the feature's dimension to 50.

%%%%%%%%%%%%%%%%%%%%%%%%%%%%%%%%%%%%%%%%%%%%%%%%%%%%%%%%%%%%%%%%%%%%%%%%%%%%%%%%
\section{Experimental Setup}
\label{app:setup}

\subsection{Meta-World tasks and their true rewards}
\label{app:expdetail}

Meta-World \cite{yu2020meta} trains and tests on a single task with a fixed reset and goal; we randomize the task instantiation in all of our experiments. We use five Meta-World tasks in total: \textit{Button-press}, \textit{Sweep-into}, \textit{Door-open} and \textit{Door-unlock} form the state-based evaluation of \cref{sub:1}, \textit{Button-press} and \textit{Sweep-into} are additionally used in their image-based variants, and \textit{Window-open} is used only for the human study of \cref{sub:8}. They cover different actions of the robotic arm within the task space, ranging from simple movements to complex manipulations. In \textit{Button press}, the robot needs to press a horizontal button with its end effector and the button position is random. In \textit{Sweep into}, the robot is required to sweep a puck into a random hole. In \textit{Door open}, the robot is asked to open a door with a revolving joint, where the door is randomized. In \textit{Window open}, the robot requests to push and open a random window. In \textit{Door unlock}, the robot is required to unlock the door by rotating the lock counter-clockwise. All of them are 500 steps long. 

In state-based tasks, the robot receives 39-dimensional states, including: a 6-tuple of the 3D Cartesian positions of the end-effector, a normalized measurement of how open the gripper is, the 3D position of the first object, the quaternion of the first object, the 3D position of the second object, the quaternion of the second object, all of the previous measurements in the environment, and finally the 3D position of the goal. In our experiments, there is no second object, so the quantities corresponding to them are zeroes. 

In image-based tasks, the robot perceives $64\times64\times3$ images from a third-person camera. Meanwhile, it can obtain its proprioceptive states, containing the 3D Cartesian positions of the end-effector and a normalized measurement representing how
open the gripper is. We also assume that the robot knows the initial position of the object and the position of the goal, which can be used to compute proxy rewards.

We provide the form of the environment's true reward. Specifically, we first define the following variables:

\begin{itemize}
    \item Object position: $o\in \mathbb{R}^3$
    \item End-effector position: $h\in \mathbb{R}^3$
    \item Target position: $t\in \mathbb{R}^3$
    \item Initial position of the object: $o_i\in \mathbb{R}^3$
    \item Initial position of the end-effector: $h_i\in \mathbb{R}^3$
    \item Gripper close/open amount: $c\in [0,1]$ (we reserve $g$ for the goal, as in \cref{tab:priors})
\end{itemize}

The tolerance function is used frequently in defining true rewards. It equals $1$ inside $[b_{min},b_{max}]$ and decays outside it at a rate set by the margin $m$; the second argument of $S$ is the value it takes one margin away from the nearest bound, which Meta-World fixes at $0.1$. Meta-World's default sigmoid is the long-tail $S$ below; \cref{app:dmc} uses two others.
$$  L(x,b_{min},b_{max},m)=\left\{
\begin{array}{ll}
      1 & b_{min}\leq x\leq b_{max} \\
      S\left(\frac{b_{min}-x}{m}, 0.1\right) & x<b_{min} \\
      S\left(\frac{x-b_{max}}{m}, 0.1\right) & x\geq b_{max},
\end{array} \right.$$
$$\text{where} \ \ S(a_1, a_2)=\left(\left(\frac{1}{a_2} - 1\right)a_1^2 + 1\right)^{-1}.$$
We use $T_{H_0}$ to represent the Hamacher product.

\paragraph{Button press} The reward function incentivizes the agent to approach and press the button. It consists of two key components whose product constitutes the reward function: (1) proximity reward, encouraging the TCP to get close to the button, and (2) button press reward, rewarding the agent when the button is sufficiently pressed. The specific form of the reward is as follows:
$$R=\left\{
\begin{array}{ll}
      2T_{H_0}(
        c,
        L(\lVert o - h \rVert, 0, 0.05, \lVert o - h_i \rVert))
        & \lVert o - h \rVert > 0.05 \\
      2T_{H_0}(
        c,
        L(\lVert o - h \rVert, 0, 0.05, \lVert o - h_i \rVert)) \\ +
        8L(| t_{(y)} - o_{(y)} |, 0, 0.005, | t_{(y)} - o_{i,(y)} |))
        & otherwise.
\end{array} \right.$$
\paragraph{Sweep into} The reward function encourages the agent to grasp the object and move it to the target location. It consists of two main components: (1) grasping reward, which measures how well the gripper cages the object, and (2) in-place reward, which evaluates how close the object is to the target position. The specific form of the reward is as follows:
$$R=\left\{
    \begin{array}{ll}
    \begin{aligned}
        & (2R_{cage,dense}(0.02,0.05,0.01) \\
        & + 2T_{H_0}\left(
            R_{cage,dense}(0.02,0.05,0.01),
            L(\lVert t - o \rVert, 0, 0.05, \lVert t - o_i \rVert)
        \right))
    \end{aligned}
            & \lVert t - o \rVert > 0.05 \\
        10
            & otherwise,
    \end{array}
\right.$$
$$\text{where} \ \  R_{cage,dense}(c_1, c_2, c_3)=\left\{
\begin{array}{ll}
      0.5(C + T_{H_0}(C, c)) & C > 0.97 \\
      0.5C & otherwise,
\end{array} \right.$$
$$C(c_1, c_2, c_3)=T_{H_0}(T_{H_0}(C_{LR,(0)}, C_{LR,(1)}), C_P(c_3)),$$
$$\text{and} \ \ C_P(c_3) = L\left(
\lVert o_{(xz)} - h_{(xz)} \rVert,
0,
c_3,
\lVert o_{i,(xz)} - h_{i,(xz)} \rVert - c_3
\right), $$ $$ C_{LR}(c_1, c_2) = L\left(
\left| \begin{bmatrix} h_{L,(y)} \\ h_{R,(y)} \end{bmatrix} - o_{(y)}\right|,
c_1,
c_2,
\left| \left| \begin{bmatrix} h_{L,(y)} \\ h_{R,(y)} \end{bmatrix} - o_{i,(y)}\right| - c_2\right|
\right).$$
\paragraph{Door open} The grasping reward is used in addition to an opening reward, which measures progress toward fully opening the door. The specific form of the reward is as follows:
$$alt = \mathbb{I}_{\lVert h_{(xy)} - o_{(xy)} \lVert > 0.12} \cdot \left(0.4 + 0.04\log\left(\lVert h_{(xy)} - o_{(xy)} \lVert-0.12\right) \right),$$
$$ready=\left\{
\begin{array}{ll}
\begin{aligned}
& T_{H_0}(
    L(\lVert h-o-\langle0.05,0.03,-0.01\rangle \rVert, 0, 0.06, 0.5), \\
    & L(alt-h_{(z)}, 0, 0.01, \nicefrac{alt}{2}),
) \\
\end{aligned}
& h_{(z)} < alt \\

L(\lVert h-o-\langle0.05,0.03,-0.01\rangle \rVert, 0, 0.06, 0.5)
& otherwise,
\end{array} \right.$$
$$R=\left\{
\begin{array}{ll}
2T_{H_0}\left(c, ready\right) +
8\left( 0.2\mathbb{I}_{o_{(\theta)} < 0.03} + 0.8L(o_{(\theta)} + \frac{2\pi}{3}, 0, 0.5, \frac{\pi}{3}) \right)
& |t_{(x)} - o_{(x)}| > 0.08 \\
10 & otherwise.
\end{array} \right.$$
\paragraph{Window open} It uses both a (1) reach reward, which encourages the agent to move its end-effector close to the handle, and (2) in-place reward, which rewards the agent for successfully pulling the handle to the open position. The specific form of the reward is as follows:
$$R=10T_{H_0}(
    L(| t_{(x)} - o_{(x)} |, 0, 0.05, | t_{(x)} - o_{i,(x)} |),
    L(\lVert o - h \rVert, 0, 0.02, \lVert o_i - h_i \rVert - 0.02).
)$$
\paragraph{Door unlock} The two reward components are (1) ready-to-push reward, which encourages the agent to align its gripper with the lock, and (2) unlock reward, which is maximized when the lock is fully pushed to the target position. The specific form of the reward is as follows:
$$R=
\begin{aligned}
2L(&\lVert \langle1,4,2\rangle \cdot (o - h + \langle0, 0.055, 0.07\rangle) \rVert, 
0, 0.02,\\
&\lVert \langle1,4,2\rangle \cdot (o_i - h_i + \langle0, 0.055, 0.07\rangle) \rVert)) + 
8L(|t_{(x)} - o_{i,(x)}|, 0, 0.005, 0.1).
\end{aligned}$$

\subsection{DM-Control tasks}
\label{app:dmc}

We use two locomotion tasks from the DeepMind Control Suite \cite{tunyasuvunakool2020}, \textit{Walker-walk} and \textit{Quadruped-walk}, to test whether RRM transfers beyond manipulation. In \textit{Walker-walk} a planar biped must reach and hold a forward speed of $1$\,m/s while keeping its torso upright; in \textit{Quadruped-walk} a four-legged robot in $3$D must hold a forward speed of $0.5$\,m/s, with a substantially higher-dimensional observation and action space that makes it far more sensitive to the initial state. \textit{Walker-walk} has a $24$-dimensional observation (fourteen body-part orientations, torso height, and nine generalized velocities) and a $6$-dimensional action; \textit{Quadruped-walk} has a $78$-dimensional observation (a $44$-dimensional egocentric state, torso velocity, torso uprightness, a six-axis IMU, and $24$ force-torque readings) and a $12$-dimensional action. Both use episodes of $1{,}000$ control steps.

We provide the form of the environment's true reward. The suite builds it from the same tolerance construction as \cref{app:expdetail}, but with a different sigmoid, a different value at the margin, and always one-sidedly ($b_{max}=\infty$). We therefore write
$$L_{\star,\kappa}(x,b,m)=\left\{
\begin{array}{ll}
      1 & x\geq b \\
      S_{\star}\!\left(\frac{b-x}{m},\, \kappa\right) & x < b,
\end{array} \right.$$
$$\text{where} \ \ S_{\mathcal{N}}(a_1,a_2)=\exp\!\left(-\tfrac{1}{2}\left(a_1\sqrt{-2\ln a_2}\right)^{2}\right) \ \ \text{and} \ \ S_{\text{lin}}(a_1,a_2)=\max\!\left(0,\ 1-(1-a_2)\,a_1\right),$$
so that $\kappa$ is the value the function takes one margin away from the bound and $\star$ names the sigmoid. In this notation the $L$ of \cref{app:expdetail} is $L_{\text{lt},0.1}$ with the long-tail $S$ defined there, whereas the suite defaults to $S_{\mathcal{N}}$ and uses $S_{\text{lin}}$ where a task requests it. We write $h_{\text{torso}}$ for the torso height, $v_x$ for the horizontal velocity of the center of mass, and $z_{\text{torso}}$ for the vertical component of the torso's own $z$-axis, so that $z_{\text{torso}}=1$ is perfectly upright and $z_{\text{torso}}=-1$ inverted.

\paragraph{Walker walk} The reward is the product of two components: (1) a stand reward, itself a weighted combination of a height term and an uprightness term, and (2) a move reward that grows with the horizontal velocity. The specific form of the reward is as follows:
$$R_{\text{stand}} = \frac{3\,L_{\mathcal{N},0.1}(h_{\text{torso}},\, 1.2,\, 0.6) + \frac{1+z_{\text{torso}}}{2}}{4},$$
$$R_{\text{move}} = L_{\text{lin},0.5}(v_x,\, 1,\, 0.5) = \min\left(\max(v_x, 0),\, 1\right),$$
$$R = R_{\text{stand}} \cdot \frac{5R_{\text{move}} + 1}{6}.$$
\paragraph{Quadruped walk} The two components are (1) an uprightness reward, which is maximized when the torso's own $z$-axis is aligned with the world's, and (2) a move reward on the forward torso velocity. They are again combined multiplicatively. The specific form of the reward is as follows:
$$R_{\text{upright}} = L_{\text{lin},0}(z_{\text{torso}},\, 1,\, 2) = \frac{1+z_{\text{torso}}}{2},$$
$$R_{\text{move}} = L_{\text{lin},0.5}(v_x,\, 0.5,\, 0.5) = \min\left(\max(v_x + 0.5, 0),\, 1\right),$$
$$R = R_{\text{upright}} \cdot R_{\text{move}}.$$

The move reward saturates at the target speed instead of peaking there, so moving faster than required is never penalized; and posture and locomotion enter as a product, so neither can be traded against the other, because a walker that has fallen earns nothing however fast it slides.

Both rewards multiply an uprightness term by a speed term, and only the uprightness term can be written down without knowing the target speed. Our prior therefore keeps that factor alone, as the upright-torso penalty
\begin{equation}
r^0_{\text{DMC}}(s) = \frac{z_{\text{torso}}(s) - 1}{2} \in [-1, 0],
\end{equation}
which is $0$ when the torso is perfectly upright and $-1$ when it is inverted. It is directly observable in both tasks: $z_{\text{torso}}$ is an explicit field of the \textit{Quadruped-walk} observation, and for the planar \textit{Walker-walk} it equals the first body orientation entry. The prior thus encodes ``stay upright'', leaving the residual to supply the entire notion of forward progress, which is why RRM still needs preferences here even though the prior alone is enough to keep the robot standing.

Because a single episode carries far more reward signal than a Meta-World episode, we query preferences much less frequently: 50 preferences every 30{,}000 steps for \textit{Quadruped-walk} (20 sessions, 1{,}000 queries in total) and every 20{,}000 steps for \textit{Walker-walk} (10 sessions, 500 queries in total). We report the true accumulated return rather than a success rate, since these tasks have no binary success criterion.

\subsection{Prior reward construction}
\label{app:priors}

\paragraph{Normalization} The residual $r'_\psi$ is bounded to $[-1,1]$ by its terminal \texttt{tanh}, so a prior whose spread is far larger would be uncorrectable by the residual, while a prior far smaller would be drowned out by it. We guard against the first case with a light, scale-only normalization,
\begin{equation}\label{eq:norm}
N(x) = \frac{x}{\max\left(1, \hat{\sigma}\right)},
\end{equation}
where $\hat{\sigma}$ is a running estimate of the standard deviation of the raw prior value over the transitions seen so far. Two properties matter. First, $N$ rescales but never shifts, so it cannot change which of two segments the Bradley-Terry model prefers. Second, the floor at $1$ makes $N$ exactly the identity whenever the prior is already on the residual's scale. The proxy priors are negative Euclidean distances and are of order $10^{-2}$ to $1$, the language priors are mapped onto $[-1,1]$, the image-based penalty proxy takes values in $\{-1,0\}$, the DM-Control upright penalty lies in $[-1,0]$, and the imitation prior is the bounded output of a discriminator. $N$ is therefore a safeguard for priors supplied by a user rather than a tuned component of the method, and it leaves all reported experiments unchanged. The replay buffer stores $r^0$ alongside $\hat{r}^{\text{RRM}}_\psi$ (\cref{alg:RRM}) so that relabeling after a reward-model update reuses exactly the prior value that was seen at collection time.

\paragraph{Proxy rewards} Proxy rewards are hand-written heuristics of the form $r(s,a) = \sum_i w_i \phi_i(s,a)$. \cref{tab:priors} lists every proxy variant that appears in our figures together with its definition, so that the legend entries can be read unambiguously.

\begin{table}[ht]
\centering
\caption{\textbf{Prior rewards used in this paper.} $s_{\text{ee}}$ is the end-effector position, $s_{\text{obj}}$ the object position, $g$ the goal, and $N(\cdot)$ the normalization defined above. The residual is always a learned network; the table describes only the prior term $r^0$.}
\label{tab:priors}
\resizebox{\linewidth}{!}{
\begin{tabular}{lll}
\toprule
\textbf{Name in figures} & \textbf{Setting} & \textbf{Definition of $r^0$} \\
\midrule
proxy            & Meta-World, state & $N(-\Vert s_{\text{obj}} - g\Vert - \Vert s_{\text{ee}} - s_{\text{obj}}\Vert)$ \\
proxy o2g        & Meta-World, state & $N(-\Vert s_{\text{obj}} - g\Vert)$, object-to-goal term only \\
proxy h2o        & Meta-World, state & $N(-\Vert s_{\text{ee}} - s_{\text{obj}}\Vert)$, hand-to-object term only \\
sparse proxy     & Meta-World, state & $1$ if the task-success condition holds, and $0$ otherwise \\
zero proxy       & Meta-World, state & $r^0 \equiv 0$ (RRM degenerates to a plain preference reward) \\
opposite proxy   & Meta-World, state & $-\,$proxy, i.e.\ the negated heuristic \\
random proxy     & Meta-World, state & a random value at the same scale as the residual \\
proxy            & DM-Control        & $(z_{\text{torso}} - 1)/2 \in [-1,0]$, an upright-torso penalty \\
penalty proxy    & Meta-World, image & $-1$ if $s_{\text{ee}} \notin \mathcal{F}$, and $0$ otherwise \\
1D proxy         & sim-to-real, Reach & $-\Vert s_{\text{ee},y} - s_{\text{goal},y}\Vert$ \\
proxy h2o        & sim-to-real, Push / Pick-and-Reach & $-\Vert s_{\text{obj}} - s_{\text{ee}}\Vert$ \\
\midrule
language         & all               & code generated by GPT-5, see below \\
imitation        & all               & reward extracted from a pre-trained AIRL model, see below \\
\bottomrule
\end{tabular}
}
\end{table}

In the image-based setting the object state is unavailable, so $\mathcal{F}$ is built only from quantities the agent can actually observe. It is the union of two rectangular cuboids of width $0.2$\,m, one centerd on the segment from the end-effector's initial position to the object's initial position and one centerd on the segment from the object's initial position to the target; membership is tested with a convex-hull check at every step. The proxy therefore penalizes leaving a corridor that a successful trajectory would plausibly stay inside, without ever referring to where the object currently is.

\paragraph{Language rewards} Language priors are produced by prompting GPT-5 \cite{gpt5} once per task, in the spirit of Eureka \cite{ma2024eureka}. The prompt has three parts: (i) a one-line statement of the benchmark and of the task in natural language, (ii) the environment's source file pasted verbatim, and (iii) an instruction to design a reward for that task. Pasting the source rather than a hand-written summary is deliberate: it exposes the observation fields, the physics accessors and the task constants exactly as the environment defines them, so the choice of what matters is left entirely to the model. For \textit{Walker-walk} the prompt is

\begin{quote}\footnotesize\ttfamily
Here is a DM-control (built on MuJoCo) task: Walker Walk. The robot Walker is required to walk normally. The environment code is:
\end{quote}
\begin{quote}\footnotesize
\begin{verbatim}
"""Planar Walker Domain."""
... (imports)

_DEFAULT_TIME_LIMIT = 25
_CONTROL_TIMESTEP   = .025
_STAND_HEIGHT       = 1.2
_WALK_SPEED         = 1
_RUN_SPEED          = 8

... (SUITE.add task constructors: stand / walk / run)

class Physics(mujoco.Physics):
  def torso_upright(self):
    """Returns projection from z-axes of torso to the z-axes of world."""
    return self.named.data.xmat['torso', 'zz']
  def torso_height(self):
    """Returns the height of the torso."""
    return self.named.data.xpos['torso', 'z']
  def horizontal_velocity(self):
    """Returns the horizontal velocity of the center-of-mass."""
    return self.named.data.sensordata['torso_subtreelinvel'][0]
  def orientations(self):
    """Returns planar orientations of all bodies."""
    return self.named.data.xmat[1:, ['xx', 'xz']].ravel()

class PlanarWalker(base.Task):
  ... (__init__, initialize_episode)
  def get_observation(self, physics):
    """Returns an observation of body orientations, height and velocites."""
    obs = collections.OrderedDict()
    obs['orientations'] = physics.orientations()
    obs['height']       = physics.torso_height()
    obs['velocity']     = physics.velocity()
    return obs
\end{verbatim}
\end{quote}
\begin{quote}\footnotesize\ttfamily
Please design a reward for this task to enable the agent to achieve the goal.
\end{quote}

Note that the suite's own reward is never shown; the constants \texttt{\_STAND\_HEIGHT} and \texttt{\_WALK\_SPEED} and the four physics accessors are the only hints the model gets, and it has to decide by itself which of them the task depends on. The prompt also does not pin down an output format, so a returned reward has to be transcribed into an executable function before it can be used, which is why we sample five candidates and keep the best-executing one. Appending a short code skeleton to part (iii) and asking the model to fill it in would remove that step. The selected function is then frozen: unlike Eureka \cite{ma2024eureka}, we never re-prompt the model or iterate on the reward using training feedback, because the point of the experiment is to test whether a single, un-tuned language reward is already a useful prior.

For the \textit{Walker-walk} prompt above, GPT-5 returned the following (the rest of the returned class reproduces the prompt verbatim and is omitted):

\begin{quote}\footnotesize
\begin{verbatim}
_WALK_SPEED = 1.0
_WALK_SIGMA = 0.25            # tolerance for Gaussian shaping

def _gaussian(x, mu, sigma):
    return np.exp(-0.5 * ((x - mu) / sigma) ** 2)

def get_reward(self, physics):
    # --- 1. Forward velocity tracking ---
    r_velocity = _gaussian(physics.horizontal_velocity(),
                           self._move_speed, _WALK_SIGMA)
    # --- 2. Upright posture reward ---
    r_upright  = (physics.torso_upright() + 1) / 2.0
    # --- 3. Height reward (stay near stand height) ---
    r_height   = _gaussian(physics.torso_height(), _STAND_HEIGHT, 0.2)
    # --- Final reward as weighted sum ---
    return float(1.0 * r_velocity + 0.5 * r_upright + 0.5 * r_height)
\end{verbatim}
\end{quote}

The model identified the right three quantities, i.e., forward velocity, uprightness and torso height, but combined them in a way that differs from the suite reward of \cref{app:dmc} in three respects. First, it takes a weighted sum where the suite takes a product. It makes the two posture terms contribute up to $1.0$ out of a maximum of $2.0$, so a walker that simply stands still at the right height already collects half the available reward, which the multiplicative form denies it. Second, both shaping terms are two-sided Gaussians, whereas the suite uses one-sided tolerances with bounds $(\cdot,\infty)$. The generated reward therefore penalizes walking faster than $1$\,m/s and standing taller than $1.2$\,m, neither of which the task treats as a mistake. Last, $\sigma = 0.25$ makes the velocity term nearly flat at low speed. In early training, when the walker is barely moving, the only usable gradient comes from the two posture terms, i.e., from exactly the stand-still optimum generated by the first difference.

This makes the consequence for RRM concrete. The language prior here is not only uninformative about locomotion, but it also actively rewards not locomoting, so the residual must first cancel part of the prior before it can contribute anything. That is a worse starting point than our upright-torso proxy, which does not encourage forward motion but also against it, and it is why the language prior converges more slowly than the proxy on DM-Control (\cref{sub:1}) even though it is the strongest of the three priors on Meta-World. We do not reproduce the functions generated for the remaining tasks: they are sampled outputs rather than a component of the method, and the prompt above is what determines them.

Writing $h$ for the end-effector position, $g$ for the goal, $h_{\text{init}}$ for the initial end-effector position and $o_{\text{init}}$ for the object's initial position, the functions selected for the two image-based tasks are
\begin{align}
r^0_{\text{lang}} &= 2\,\mathrm{clip}\!\left(1 - \frac{\Vert h - g\Vert}{\Vert g - h_{\text{init}}\Vert + \epsilon},\, 0,\, 1\right) - 1 && \text{(\textit{Button-press})}, \\
r^0_{\text{lang}} &= 2\,\mathrm{clip}\!\left(0.1\left(1 - \frac{\Vert h - g\Vert}{\Vert g - h_{\text{init}}\Vert + \epsilon}\right) + 0.9\left(1 - \frac{\Vert h - o_{\text{init}}\Vert}{\Vert o_{\text{init}} - h_{\text{init}}\Vert + \epsilon}\right),\, 0,\, 1\right) - 1 && \text{(\textit{Sweep-into})},
\end{align}
with $\epsilon = 10^{-8}$. Both refer only to the goal and to initial positions and never to the object's current pose, which is not recoverable from pixels, and both are bounded to $[-1,1]$ by construction, i.e.\ on the same scale as the residual.

\paragraph{Imitation rewards} Imitation priors are the discriminator-derived reward of Adversarial Inverse Reinforcement Learning (AIRL) \cite{fu2017airl}, trained before RRM training begins and then frozen. Demonstrations are collected by first training a SAC agent on the environment's true reward and then rolling out its deterministic policy: 50 episodes for every state-based task, and, in the image-based setting, 100 episodes for \textit{Button-press} and 500 for \textit{Sweep-into}. \textit{Sweep-into} needs the larger set because it is the harder of the two, so a fixed number of episodes covers a much smaller fraction of the relevant state space. AIRL is then trained for $10^6$ environment steps after $5{,}000$ seed steps, updating the discriminator with $100$ gradient steps every $1{,}000$ environment steps on minibatches of $128$ expert and $128$ policy transitions. Its reward and value heads are three-layer $256$-unit MLPs optimized with Adam ($\beta = (0.9, 0.999)$) at learning rate $10^{-4}$ in the state-based setting and $5\times10^{-5}$ in the image-based setting, where they read a $50$-dimensional visual feature concatenated with the $4$-dimensional proprioceptive state. We keep the $10^6$-step checkpoint at $100\%$ evaluation success as the frozen prior. In the image-based setting the AIRL discriminator necessarily contains its own convolutional encoder; we transfer that encoder directly into the policy-learning phase, since the reward model would otherwise be reading a representation that the agent does not share. \cref{app:image} reports a control experiment in which the same pre-trained encoder is given to every other method, to confirm that the gain does not only come from this transfer.

\subsection{Sim-to-real environments}
\label{app:real}

The Reach, Push and Pick-and-Reach tasks are implemented in PyBullet \cite{coumans2021} with a Franka Panda arm, mirroring the physical setup of \cref{fig:real-a} \cite{ablett2024vpace}. \cref{tab:spacesenv} gives the action and observation spaces of each task; the true rewards used to generate synthetic preferences and the proxy priors are stated in \cref{sub:6}, and training uses the Meta-World hyperparameters of \cref{tab:hparams-rrm}. For real-world evaluation we transfer the simulation policy zero-shot and estimate the cube pose from an ArUco marker observed by two calibrated Intel RealSense D435 cameras. \cref{tab:realiqm} reports the resulting real-robot performance: each entry averages 5 seeds $\times$ 20 physical episodes, with the cube and the goal re-randomized within the workspace at every episode. A trial counts as a success if the center of the cube ends within $5$\,cm of the center of the goal; for Reach, where the cube itself is the target and no separate goal is part of the observation (\cref{tab:spacesenv}), the same $5$\,cm threshold is applied between the end-effector and the cube.

\begin{table}[t]
    \centering
    \caption{\textbf{Sim-to-real environments spaces.} The action space and observation space are composed of these vectors on 3 tasks in the real world.}
    \resizebox{\textwidth}{!}{
    \begin{tabular}{llll}
        \toprule
        \textbf{Attribute}        & \textbf{Reach}           & \textbf{Push}   & \textbf{Pick-and-Reach}           \\
        \midrule
        Action Space              & X, Y, Z end-effector velocity    & X, Y end-effector velocity  & X, Y, Z end-effector velocity  \\ \cmidrule(lr){2-4}
        \multirow{3}{*}{Observation Space}        & X, Y, Z end-effector position & X, Y end-effector position & X, Y, Z end-effector position \\
        & X, Y, Z cube position & X, Y cube position and yaw & X, Y, Z cube position and yaw \\
        & & X, Y goal position & X, Y, Z goal position \\
        \bottomrule
    \end{tabular}%
    }
    \label{tab:spacesenv}
\end{table}

\subsection{Noisy teacher models}
\label{app:stoc}

We follow Lee et al.~\cite{lee2021bpref} for the two noisy teachers used in \cref{sub:3}. To get stochastic feedback, we use a stochastic model (Bradley-Terry model) to generate feedback:

\begin{equation}
    P[\sigma_0 \succ \sigma_1] = \frac{\exp(\sum_tr(s_t^1, a_t^1))}{\sum_{i\in\{0,1\}}\exp(\sum_tr(s_t^i, a_t^i))},
\end{equation}

where $r(s,a)$ is the true reward. It can be interpreted that the labels of pairs of segments are exponentially proportional to the sum over the segment of the true reward. However, it only labels according to a certain probability, not completely according to the cumulative real reward size of the two segments. For mistaken feedback, we take an otherwise deterministic teacher and flip its label independently with probability $p$; we report $p = 0.1$ and $p = 0.15$ in \cref{tab:wrongiqm1}. The two models are qualitatively different: a stochastic teacher is unbiased but noisy on close comparisons, whereas a mistaken teacher corrupts easy and hard comparisons alike and therefore injects labels that no reward function can explain.

\subsection{Human study interface and protocol}
\label{app:human}

\begin{figure}[t]
    \centering
    \begin{minipage}{0.57\textwidth}
        \centering
        \includegraphics[width=\textwidth]{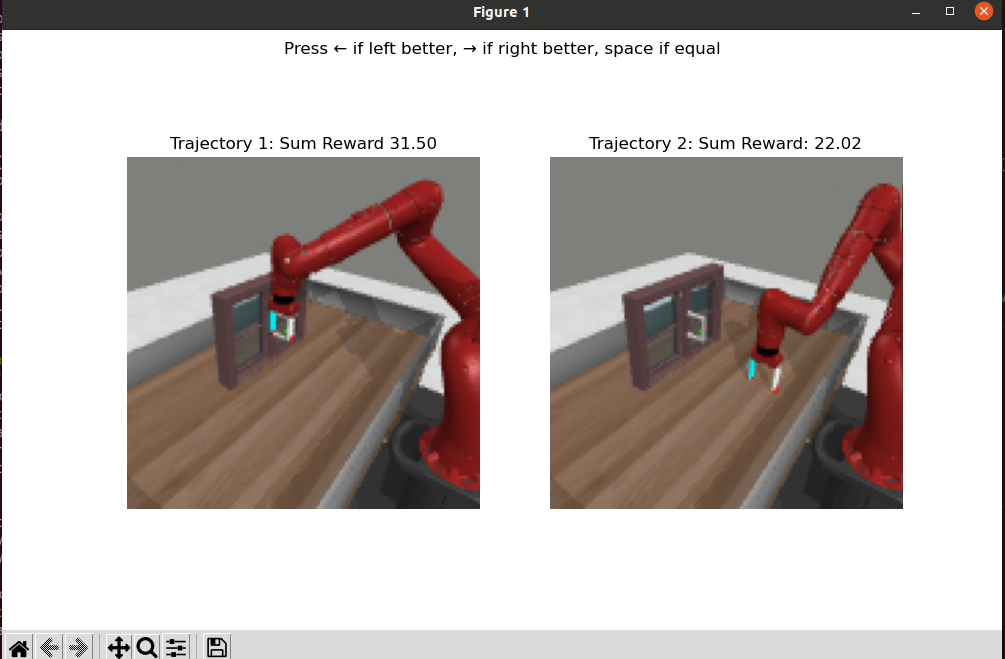}
        \subcaption{An illustration of the interface.}
        \label{fig:human-a}
    \end{minipage}
    \begin{minipage}{0.4\textwidth}
        \centering
        \includegraphics[width=\textwidth]{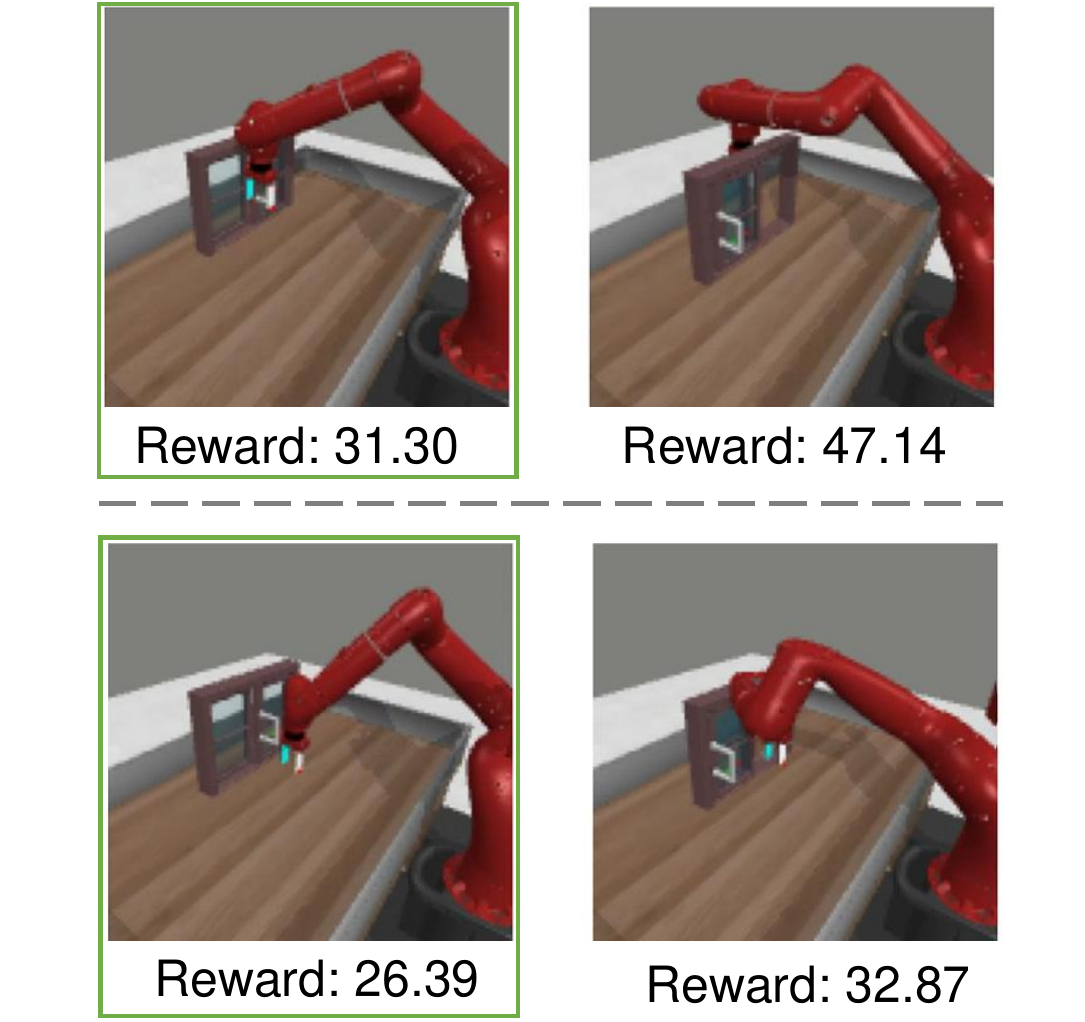}
        \subcaption{Reward comparisons.}
        \label{fig:human-b}
    \end{minipage}
    \caption{\textbf{Real-user study.} A simple interface is made for collecting feedback from real humans. Users need to press the left arrow or the right arrow on the keyboard to choose their performance. They should try to avoid giving equal feedback based on pressing the space bar, so the sum reward is helpful for reference when they don't know which one is better.}
    \label{fig:human}
\end{figure}

\cref{fig:human-a} shows the interface used to collect the real human preferences of \cref{sub:8}. Five volunteers labeled the queries; each was given the task description and a short practice round before the session began, and no volunteer labeled the same run twice. The cumulative return of each segment is displayed above the animations purely as a reference for ambiguous pairs. \cref{fig:human-b} illustrates why this does not simply turn the human into a return oracle: in comparisons of this kind, most labellers in fact preferred the segment with the lower cumulative return, because a segment that scores well under the true reward can still look clumsy or indirect to a person watching it. This is the same gap between return and human judgement that motivates preference learning in the first place, and it is why the reference number is shown as a tiebreaker rather than as the answer.

\subsection{Implementation and hyperparameters}
\label{app:hyper}

We build RRM on the public PEBBLE implementation\footnote{\url{https://github.com/rll-research/BPref}}, and the image-based variants on DrQ-v2\footnote{\url{https://github.com/facebookresearch/drqv2}} \cite{yarats2021DrQv2} and DrM\footnote{\url{https://github.com/XuGW-Kevin/DrM}} \cite{xu2023drm}.

The experiments for RRM were conducted in a Python 3.8 environment with Pytorch 1.4.0. Our setup included CUDA version 11.3, running on Ubuntu 20.04. The hardware used comprised four GeForce RTX 4090 GPUs and an Intel(R) Xeon(R) Platinum 8358P CPU @ 2.60GHz.

The full list of hyperparameters is presented in \cref{tab:hparams-sac,tab:hparams-rrm,tab:hparams-drq}. Apart from the task-dependent feedback budgets listed in \cref{tab:hparams-rrm}, the settings are identical for every method and every run, so that the comparison against PEBBLE isolates the effect of the residual structure.

\begin{table}[ht]
	\caption{\textbf{Hyperparameters for SAC}}
 \vspace{2pt}
	\label{tab:hparams-sac}
	\centering
	\begin{tabular}{ll|ll}
		\toprule
		\textbf{Hyperparameter} & \textbf{Value} & \textbf{Hyperparameter} & \textbf{Value} \\ 

  \midrule
  		Optimizer      &  Adam     &  Critic hidden layers   & 3    \\
		Actor learning rate          & 1e-4   &   Critic activation function    & ReLU   \\
		Actor hidden dim            &  256   &    Critic target update freq    & 2   \\ 
		Actor hidden layers           &  3 & Critic EMA     &   0.005\\ 
		Actor activation function  &  ReLU    &  Discount   &   0.99  \\ 
		Critic learning rate      &  1e-4  &    Init temperature    &   0.1   \\
		Critic hidden dim           & 256   & Bacth size   & 512 \\
        \bottomrule
	\end{tabular}
\end{table}

\begin{table}[ht]
	\caption{\textbf{Hyperparameters for RRM}}
 \vspace{2pt}
	\label{tab:hparams-rrm}
	\centering
	\begin{tabular}{ll}
		\toprule
		\textbf{Hyperparameter} & \textbf{Value} \\
  \midrule
  		Length of segment      &  50    \\
        Unsupervised pre-training steps   & 0   \\
		 Number of queries per session (reward batch size)   &  50  \\
         Size of preference buffer    & 10000   \\
         Frequency of feedback (Meta-World)   & every 5{,}000 steps   \\
         Frequency of feedback (\textit{Quadruped-walk})   & every 30{,}000 steps   \\
         Frequency of feedback (\textit{Walker-walk})   & every 20{,}000 steps   \\
         \midrule
         \multicolumn{2}{l}{Total feedback (task-dependent)} \\
         \quad \textit{Button-press}, \textit{Door-open}, \textit{Door-unlock} (state and image)  & 5{,}000 \\
         \quad \textit{Sweep-into} (state and image)   & 10{,}000 \\
         \quad \textit{Quadruped-walk}   & 1{,}000 \\
         \quad \textit{Walker-walk}   & 500 \\
        \bottomrule
	\end{tabular}
\end{table}

\begin{table}[htbp]
	\caption{\textbf{Hyperparameters for DrQ-v2}}
 \vspace{2pt}
	\label{tab:hparams-drq}
	\centering
	\begin{tabular}{ll|ll}
		\toprule
		\textbf{Hyperparameter} & \textbf{Value} & \textbf{Hyperparameter} & \textbf{Value} \\ 

  \midrule
  		$n$-step returns      &  3     &  Batch size   & 512    \\
		Optimizer          & Adam   &   Discount    & 0.99   \\
		 Learning rate   &  1e-4   &    Agent update frequency    & 2   \\ 
         Critic soft-update rate   &  0.01   &   Exploration stddev. clip    & 50    \\ 
         Hidden dimension   &  256   &   Exploration stddev. schedule    & linear(1.0, 0.1, 500000)   \\ 
         Feature dimension    & 50 & & \\
        \bottomrule
	\end{tabular}
\end{table}

%%%%%%%%%%%%%%%%%%%%%%%%%%%%%%%%%%%%%%%%%%%%%%%%%%%%%%%%%%%%%%%%%%%%%%%%%%%%%%%%
\section{Additional Results}
\label{app:addexp}

\subsection{Why fine-tuning a reward model fails}
\label{app:ft}

\begin{figure}[ht]
  \centering
  \includegraphics[width=0.85\textwidth, trim=0 0 337pt 0, clip]{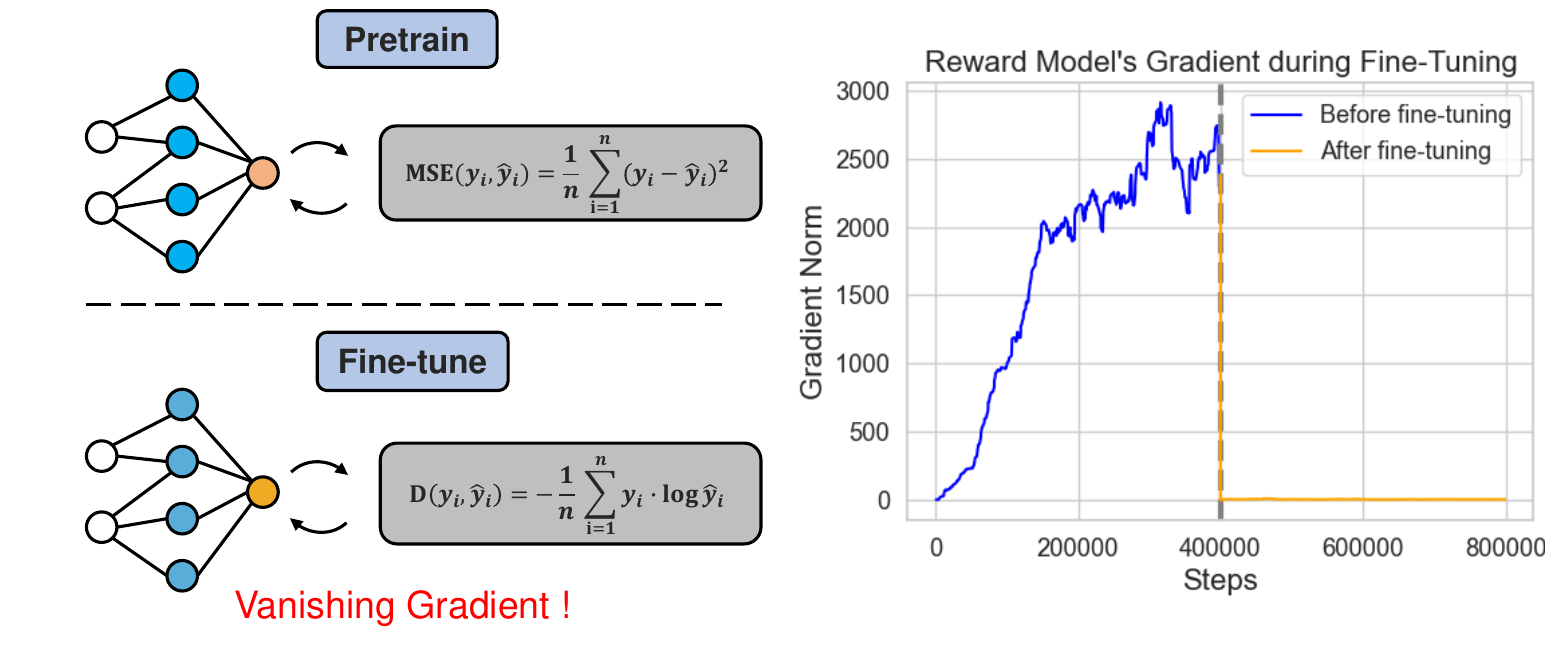}
  \caption{\textbf{Why fine-tuning a reward model is unstable.} \emph{Left:} pre-training uses an MSE loss on demonstrations while fine-tuning uses the cross-entropy loss of \cref{eq:btloss}, so the gradients seen by the reward model differ in both scale and direction at the moment of the switch. \emph{Right:} the gradient norm of a reward model pre-trained with AIRL on 50 demonstrations, measured on \textit{Button-press} across the switch to preference fine-tuning at 400{,}000 steps. It vanishes immediately after the switch, which is what stalls the fine-tuned reward in \cref{fig:ftmain}.}
  \label{fig:ft}
\end{figure}

\cref{fig:ftmain} in the main text shows that fine-tuning an imitation reward with preferences plateaus far below the RRM built on the same prior. \cref{fig:ft} explains why. Pre-training minimizes an MSE loss on demonstrations while fine-tuning minimizes the cross-entropy loss of \cref{eq:btloss}; the two objectives induce gradients of different scale and direction, so switching between them is not a smooth continuation of the same optimization problem. Measuring the gradient norm of the reward model across the switch confirms this: it collapses to near zero as soon as the loss changes at 400{,}000 steps, and the reward model stops moving. RRM avoids the switch entirely by never retraining the prior. It holds the prior and only optimizes the residual, as a single objective makes the training more stable.

\subsection{Fairness of the pre-trained encoder in image-based tasks}
\label{app:image}

\begin{figure}[t]
  \centering
  \includegraphics[width=0.8\textwidth]{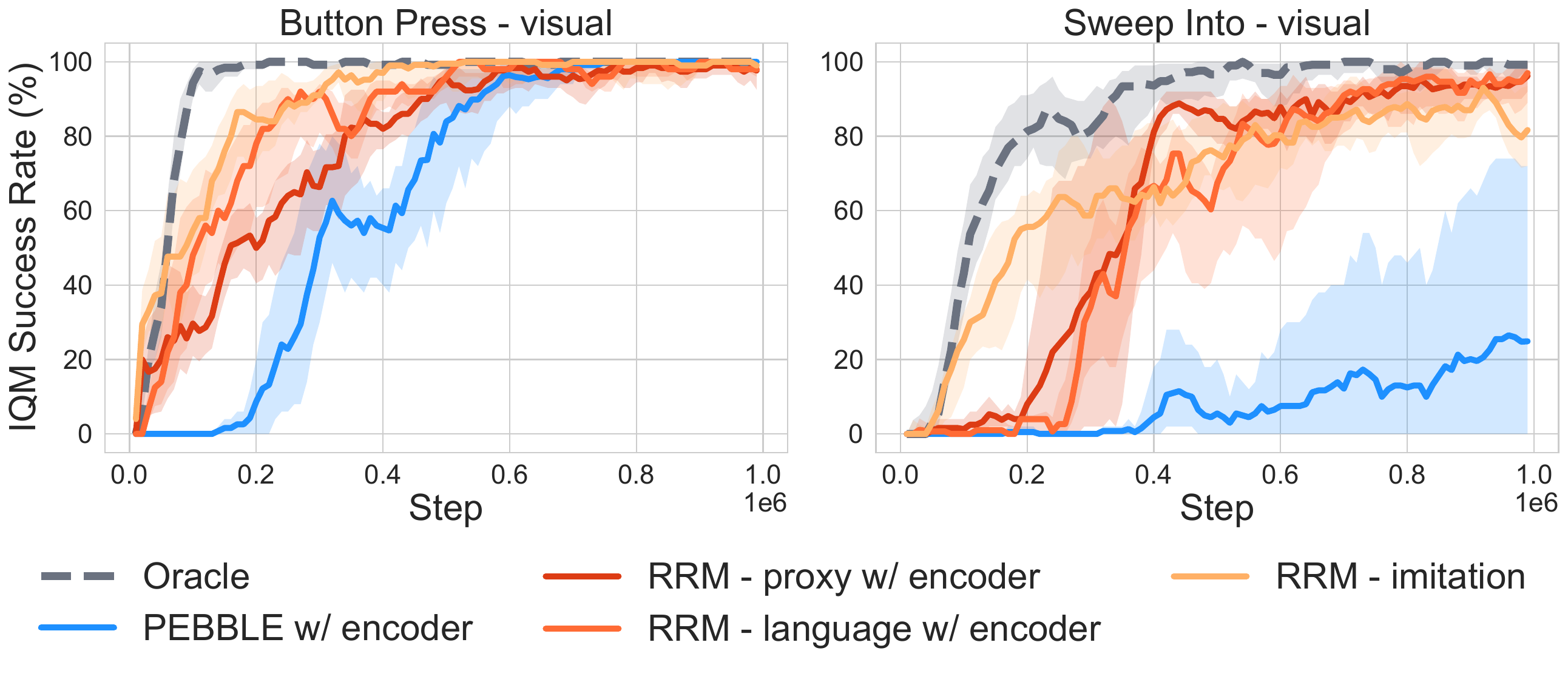}
  \caption{\textbf{Comparison with the pre-trained encoder on image-based tasks.} We apply the AIRL pre-trained encoder to PEBBLE-visual and to RRM with the proxy and language priors, so that every method starts from the same representation as RRM with the imitation prior. The oracle is indicated by a dotted line in blue.}
  \label{fig:fair_v}
\end{figure}

Since RRM with imitation reward uses the AIRL pre-trained encoder at the beginning of training to ensure the stability of the reward model, this approach may introduce some unfairness in the experiments. However, it is feasible because, before training RRM with imitation reward, we must first obtain the reward model through AIRL training. To ensure fairness, we apply this pre-trained encoder to other algorithms, as shown in \cref{fig:fair_v}. Despite adding the pre-trained encoder, PEBBLE-visual still does not perform well, especially on the \textit{Sweep-into}. However, RRM with the proxy and the language prior both show improved performance, even surpassing RRM with imitation reward. Notably, despite the addition of the encoder, RRM with imitation reward still has a faster convergence speed and maintains competitive performance. This indicates that when imitation reward can be pre-trained, it remains the preferred choice over prior rewards.

\subsection{Ablation on the residual reward in image-based tasks}
\label{app:abla}

\begin{figure}[t]
  \centering
  \includegraphics[width=0.8\textwidth]{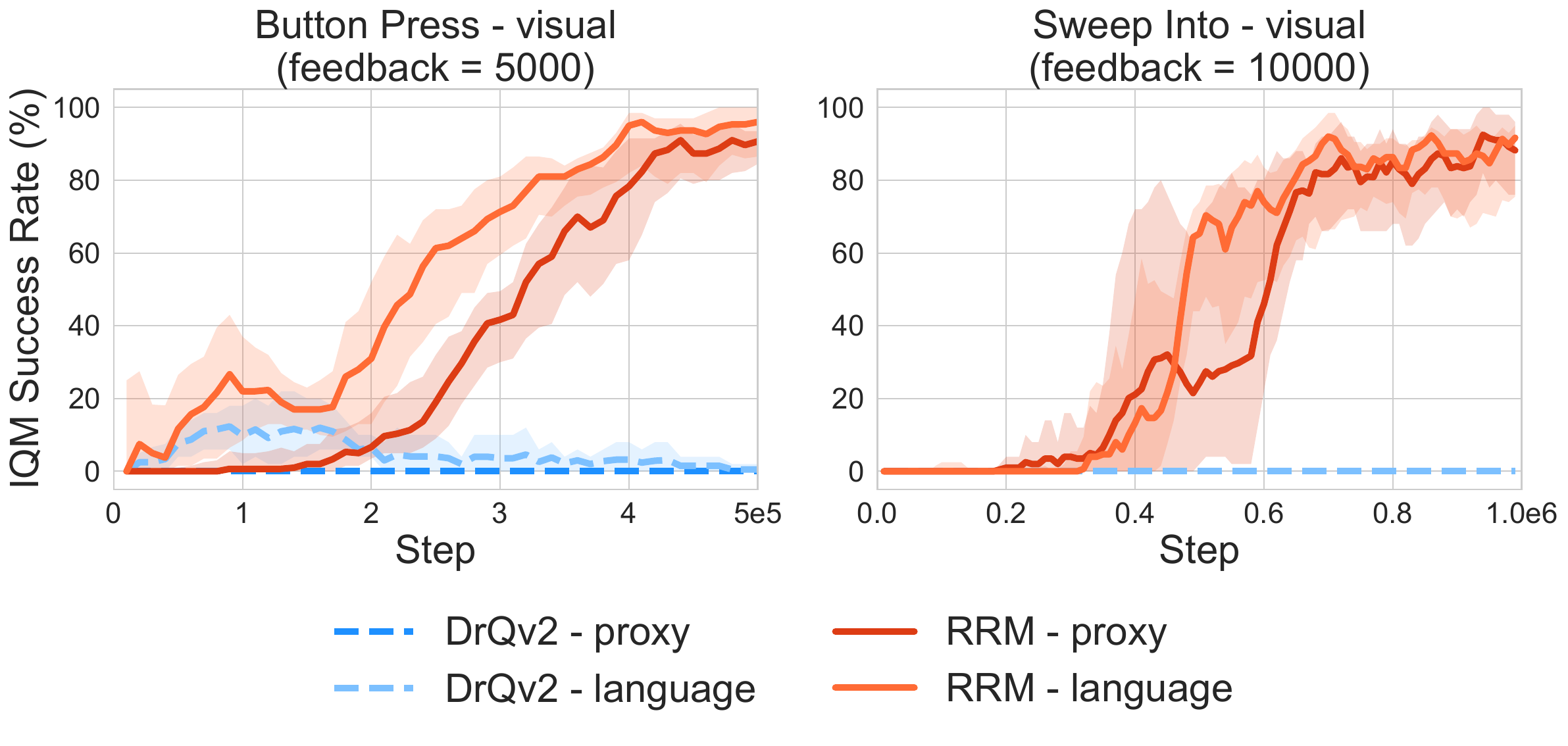}
  \caption{\textbf{Ablation on the residual reward in image-based tasks.} We compare RRM against training DrQv2 directly on the same prior, using a feedback budget of 5{,}000 queries for \textit{Button-press} and 10{,}000 for \textit{Sweep-into}, across 10 seeds.}
  \label{fig:abla_v}
\end{figure}

In image-based tasks, \cref{fig:abla_v} ablates the residual reward by training DrQv2 \cite{yarats2021DrQv2} directly on each prior. Both prior-only baselines stay near a $0\%$ success rate, whereas the corresponding RRMs solve both tasks. Without access to the object state, the penalty proxy and the language prior can only be written in terms of proprioception and the initial workspace layout, so neither of them encodes the actual task goal: an agent optimizing them learns to move through the feasible region but never completes the task. The residual reward is what supplies the missing goal information, which makes the gain from RRM larger in image-based tasks than in state-based ones.

\subsection{Why RRM tolerates an opposite prior}
\label{app:oppo}

\begin{figure}[htbp]
\centering
    \begin{minipage}{0.65\textwidth}
        \centering
        \includegraphics[width=\textwidth]{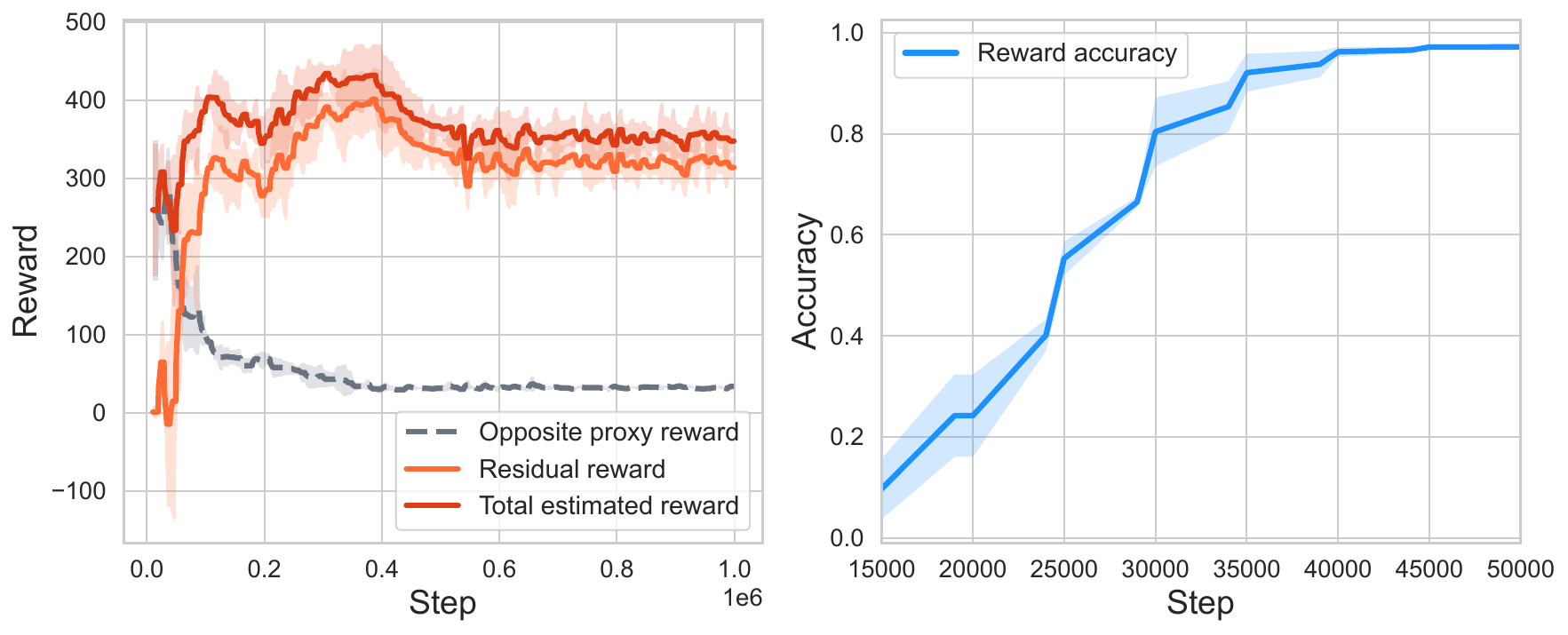}
        \subcaption{Proxy reward, residual reward and training accuracy.}
        \label{fig:oppo-a}
    \end{minipage}\hfill
    \begin{minipage}{0.33\textwidth}
        \centering
        \includegraphics[width=\textwidth]{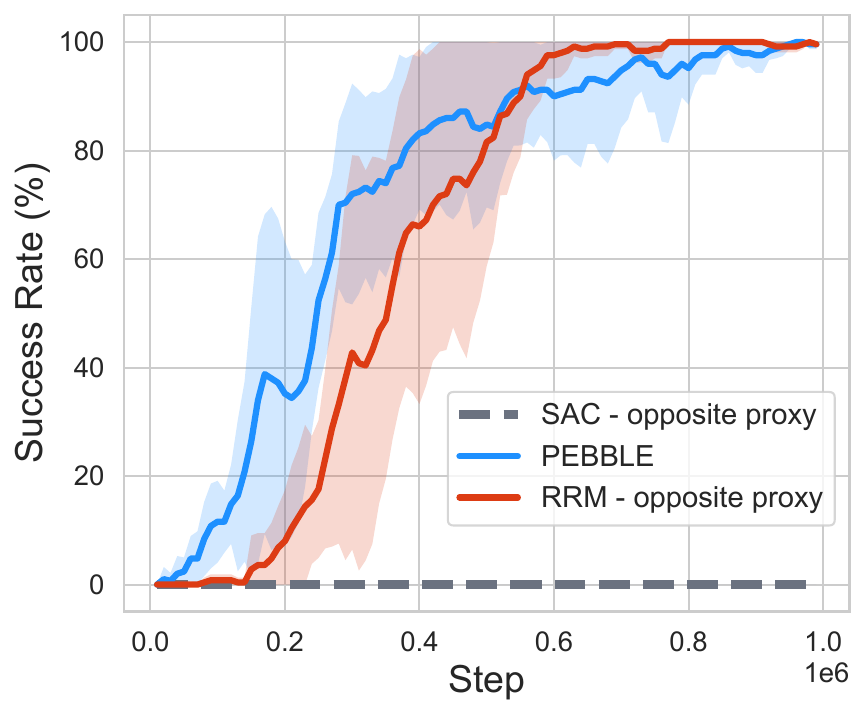}
        \subcaption{Door unlock.}
        \label{fig:oppo-b}
    \end{minipage}
  \caption{\textbf{More details on opposite reward experiments.} (a) The left plot shows the training curves of the opposite proxy reward, residual reward, and total estimated reward, where the total estimated reward is the sum of the proxy reward and the residual reward. The right plot shows the reward accuracy, which is computed as the agreement between the preferences labeled by the reward model and those derived from the true reward. (b) We further conduct an experiment on \textit{Door-unlock} for 1 million steps with 10 seeds. }
  \label{fig:oppo-app}
\end{figure}

To further explore why RRM can deal with opposite proxy reward, we report the opposite proxy reward, residual reward, total estimated reward and reward accuracy during training in \cref{fig:oppo-a}. In the early stage of training (before 30,000 steps), the total estimated reward is mainly contributed by the opposite proxy reward. However, the reward model learns quickly, showing that the reward accuracy increases to nearly 100$\%$ within 50,000 steps. That is because the opposite reward provides a strong initial signal, allowing the RRM to be ``aware'' of how to generate rewards that match the true preferences. After its convergence, the residual reward quickly becomes more dominant than the proxy reward, enabling the agent to learn a correct policy to complete the task, which in turn reduces the opposite proxy reward. 

We also conduct an experiment on \textit{Door-unlock} in \cref{fig:oppo-b} to prove that it isn't an individual case. The results show that RRM is indeed able to extract useful information from the opposite proxy reward to update the model and reduce its negative influence. However, it is still affected by the misleading message, which causes a slower convergence speed than PEBBLE.

\end{document}